\documentclass{fairmeta}

\usepackage{amsmath,amsfonts,bm}

\def\eqref#1{equation~\ref{#1}}

\def\1{\bm{1}}

\DeclareMathAlphabet{\mathsfit}{\encodingdefault}{\sfdefault}{m}{sl}
\SetMathAlphabet{\mathsfit}{bold}{\encodingdefault}{\sfdefault}{bx}{n}

\usepackage[utf8]{inputenc}
\usepackage[T1]{fontenc}
\usepackage{hyperref}
\usepackage{url}
\usepackage{booktabs}
\usepackage{graphicx}
\usepackage{amsmath}
\usepackage{amssymb}
\usepackage{nicefrac}
\usepackage{microtype}
\usepackage{xcolor}
\usepackage{tabularx}
\usepackage{array}
\usepackage[table]{xcolor}
\usepackage{siunitx}
\usepackage{xurl}
\usepackage{multirow}
\usepackage{longtable}
\usepackage{nicefrac}
\usepackage{listings}
\newcolumntype{T}{>{\raggedright\arraybackslash}X}
\newcommand{\taskname}[1]{\nolinkurl{#1}}

\title{Verify Less, Evolve More: 
Training Idea-Level Critics for Verification-Efficient ML Evolving Agents}

\author[1,\dagger,*]{Jiamu Bai}
\author[2,\dagger]{Lizhu Zhang}
\author[1,\dagger,*]{Xin Yu}
\author[2]{Yanhong Wu}
\author[2]{Zellux Wang}
\author[2]{Serena Li}
\author[2]{Weiwei Li}
\author[2]{Zhuokai Zhao}
\author[1]{Lingzhou Xue}
\author[1]{Kiwan Maeng}
\author[2]{Xiangjun Fan}
\author[2,\ddagger]{Bo Peng}

\affiliation[1]{Penn State University}
\affiliation[2]{Meta}

\contribution[*]{Work done at Meta}
\contribution[\dagger]{Equal contribution (co-first authors)}
\contribution[\ddagger]{Project Lead}

\abstract{
As large language models become more powerful, self-evolving agents are able to tackle challenging tasks including AI for machine learning (AI4ML). In AI4ML, while empirical verification is available, it often requires computationally costly model training and evaluation, limiting the speed and scale of agent evolution. Yet verification efficiency remains underexplored, and frontier models provide only limited gains when used directly as idea selectors. We address this gap with specialized \emph{idea-level critic models} that predict whether a proposed ML modification will improve upon the current solution, allowing agents to screen ideas and concentrate verification resources on the most promising candidates. We train the critic models through supervised fine-tuning on high-quality critiques synthesized by Gemini-3.1-Pro, followed by GRPO to further improve their predictive accuracy. Empirically, our critic models outperform Gemini-3.1-Pro in static idea evaluation, and these gains extend to agent inference, continual learning, and policy training. During inference-time evolution, they improve final solution quality under the same verification budget by selecting more promising ideas, with further gains from continual learning. During policy training, they serve as learned reward models, reserving empirical verification for uncertain cases and enabling substantially more policy updates with the same verification resources. Together, these results show that idea-level critic models help ML agents discover better solutions and learn stronger proposal policies under limited verification budgets.

}

\date{\today}

\begin{document}

\maketitle

\section{Introduction}
\label{section:intro}

Recent advances in large language models (LLMs) have spurred growing interest in self-improving and evolutionary agents, which iteratively propose modifications and retain successful variants based on empirical feedback~\citep{novikov2025alphaevolve,zhang2025darwingodel,sharma2025openevolve}. These efforts demonstrate unprecedented success on challenging tasks such as AI for machine learning (AI4ML), where agents develop and refine ML solutions through implementation and empirical evaluation~\citep{huang2024mlagentbench,chan2025mlebench,lu2024aiscientist}. However, AI4ML presents a distinctive verification challenge: while available, empirical verifiers often require computationally costly model training and evaluation. This execution cost accumulates across candidates and evolutionary rounds, directly limiting how many alternatives an agent can explore under a fixed compute budget.  Addressing this issue by simply using a frontier model to screen candidates offers limited relief: in our OpenEvolve experiments, directly prompting Gemini-3.1-Pro as an idea selector yields only modest gains (Section~\ref{sec:openevolve_results}). These limitations motivate us to ask: can we train specialized models to assess proposed ML modifications before empirical verification, reserving expensive execution for those most likely to improve the current solution?

\begin{figure}[t]
    \centering
    \includegraphics[width=\linewidth]{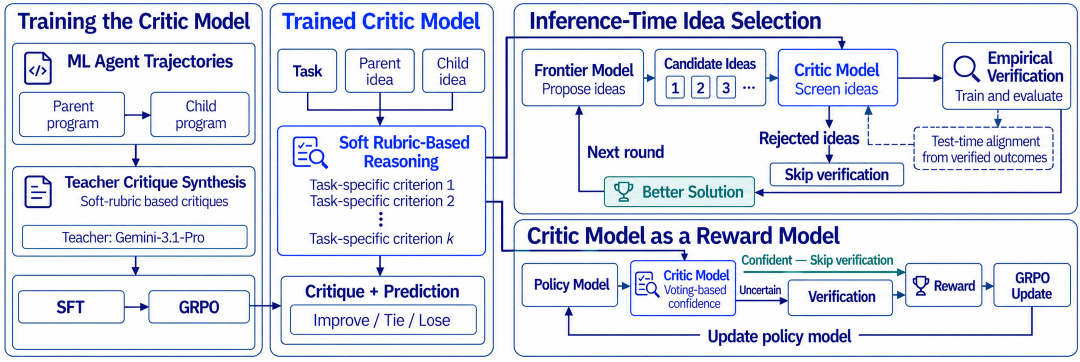}
    \vspace{-20pt}
    \caption{We train an idea-level critic model from ML agent evolution trajectories through SFT on soft-rubric critiques followed by GRPO. The trained critic model then supports (1) inference-time evolution through idea selection and task-specific test-time alignment, and (2) policy training by serving as a learned reward model while routing uncertain cases to empirical verification.}
    \vspace{-15pt}
    \label{fig:whole-pipeline}
\end{figure}

In this work, we develop a training pipeline for specialized \emph{idea-level critic models} that screens proposed ML improvements before verification, reducing reliance on computationally costly empirical verification. Given a task description, a parent solution, and a proposed child modification, the critic model predicts whether the modification will improve upon the parent. 
To train the critic model, we collect realistic idea-evolution trajectories from live ML agent runs and label parent--child pairs using their empirical execution outcomes. We then use Gemini-3.1-Pro~\citep{gemini2026gemini31pro} to synthesize critiques, retain critiques consistent with the execution-derived labels through rejection sampling, and train the critic model through supervised fine-tuning followed by GRPO~\citep{shao2024deepseekmath}. The critiques follow a soft rubric-based reasoning format: the critic model compares the parent and child under every task-specific criterion while retaining flexibility in weighting the criteria and combining their judgments. Our ablations show that this structure supports continued improvement during further RL alignment, whereas unconstrained reasoning leads to performance degradation. We hold out a portion of the collected data to form \textsc{ML-Idea-Critic-Bench}, a static benchmark for measuring training gains and comparing the idea-level assessment capabilities of different models. On this benchmark, our Qwen-3.5-9B and Qwen-3.5-35B critic models achieve three-way directional accuracies of 60.7\% and 63.3\%, respectively, substantially outperforming Gemini-3.1-Pro at 46.6\%. 

We evaluate how these predictive gains translate into improvements in agent inference, continual learning, and policy training. First, integrated into OpenEvolve~\citep{sharma2025openevolve} with Gemini-3.1-Pro as the idea proposer, the critic models screen candidate ideas before verification, improving final solution quality on MLS-Lite-Bench under matched verification budgets. Second, continual learning allows the critic models to adapt using empirical feedback accumulated during the current evolution run, further improving the final solutions. Third, while critic model improves evolution by prioritizing the most promising candidates, the quality of ideas generated remain constraint by the fixed proposer. We therefore extend the critic model from selecting among existing ideas to improving the idea-proposing policy itself.
Here, the critic model serves as a learned reward model for GRPO training of the idea-proposing policy, assigning rewards to confident predictions and routing uncertain cases to empirical verification. This reduces the real-verification cost during training, enabling 2.75$\times$ as many policy updates under the same verification budget. In summary, our main contributions are below:
\begin{itemize}
\item We establish an effective training pipeline for idea-level critic models from real ML agent evolving trajectories. Through fine-tuning from soft-based reasoning critiques, our critic model outperforms both their base models and Gemini-3.1-Pro in static idea evaluation.
\item We show that directly prompting an off-the-shelf frontier model, Gemini-3.1-Pro, as an idea selector yields only limited gains in OpenEvolve, whereas our trained critic models improve inference-time evolution under fixed verification budgets by concentrating empirical verification on the most promising proposals.
\item We extend the use of critic models from idea selection to policy training by using their predictions as rewards and reserving empirical verification for uncertain cases. This reduces verification cost and enables more policy updates under a fixed verification budget.
\end{itemize}

\begin{figure}
    \centering
    \includegraphics[width=\linewidth]{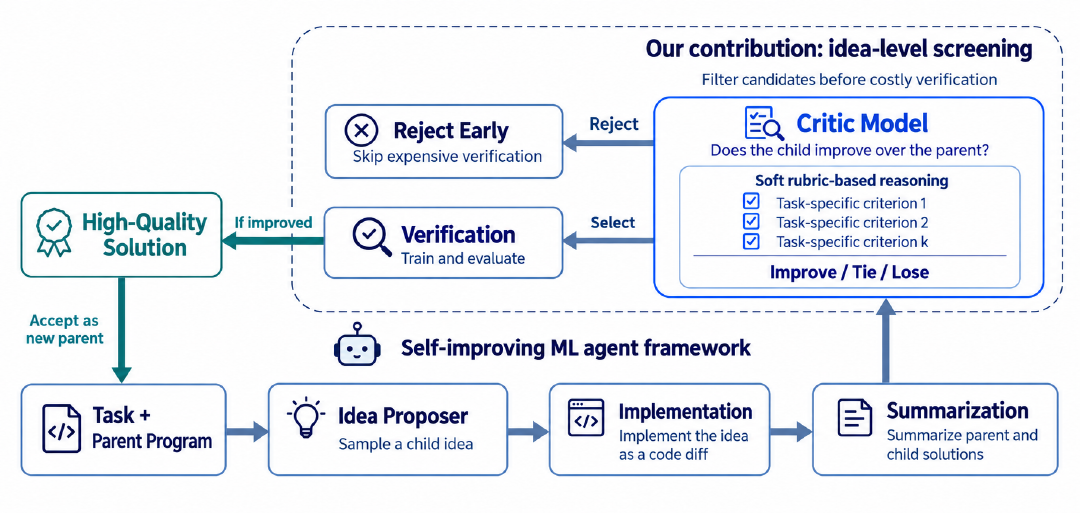}
    \vspace{-30pt}
    \caption{Overview of our idea-level screening pipeline. Given an ML task, a parent solution, and a candidate idea, the critic model predicts whether the idea will improve upon the parent. Only ideas predicted to improve are empirically verified, reducing costly verification. }
     \vspace{-15pt}
    \label{fig:framework}
\end{figure}

\section{Building Criterion-Guided Idea-Level Critic Models}
\subsection{Task Setting}
\label{sec:problem-definition}

\paragraph{Self-evolving ML agent framework.}
We consider an ML agent that iteratively improves a solution through proposal, implementation, and empirical verification. At round $r$, given a task $t$ and a parent program $P_r$, the \textbf{idea proposer} generates a modification plan $z_r$.
The \textbf{implementer} translates this plan into a child program $C_r$ while preserving the proposed modification.
An empirical evaluator $E_t$ measures solution quality, and the agent retains the better program:
\begin{equation}
\begin{aligned}
z_r &= \operatorname{Propose}(t, P_r), \qquad
C_r = \operatorname{Implement}(P_r, z_r),
\end{aligned}
\end{equation}
and the parent is updated as child if child achieves better performance,
\begin{equation}
    P_{r+1} =C_r \text{ if } E_t(C_r) > E_t(P_r) \text{ else } P_r.
\end{equation}
Repeating this loop allows the agent to accumulate improvements, but verifying each candidate can require computationally costly model training and evaluation. We show an example of such framework in Appendix~\ref{app:pipeline-example}.

\paragraph{Critic Model-Augmented ML Agent.}
Our goal is to predict the outcome of empirical verification from idea-level descriptions, enabling candidate screening before execution. To achieve this, we augment the agent pipeline with a critic model, as illustrated in Figure~\ref{fig:framework}.
To provide clear and neutral inputs, a \textbf{summarizer} converts the parent and child programs into concise descriptions $p_r$ and $a_r$ of their implemented approaches, removing exploratory reasoning, redundant details, and persuasive claims from the idea proposer's output.
At a high level, the critic model parameterized by $\theta$ uses $(t,p_r,a_r)$ to predict the execution-derived outcome:
\begin{equation}
\max_{\theta}\;
\mathbb{E}_{(t,P_r,C_r)\sim\mathcal{D}}
\left[\Pr_{\theta}\!\left(\operatorname{Critic}_{\theta}(t,p_r,a_r)=y_r\right)\right]
,
y_r =
\begin{cases}
\textsc{Improve}, & E_t(C_r) > E_t(P_r). \\
\textsc{Tie},     & E_t(C_r) = E_t(P_r). \\
\textsc{Lose},    & E_t(C_r) < E_t(P_r).
\end{cases}
\end{equation}
Here, $\mathcal{D}$ denotes the distribution of task--parent--child instances. These predictions allow the agent to prioritize promising candidates for empirical verification. 

\subsection{Criterion-Guided Critic Model Design}
\label{sec:criterion_reasoning}

To make idea-level evaluation task-aware, we structure the critic model's reasoning around a set of task-specific criteria. The critic model must compare the parent and child solutions under each criterion before reaching a final verdict. For each ML task, we use Claude-4.8-Opus~\citep{ClaudeOpus48} in a multi-round probing process to analyze the task setup, data, evaluation metric, and modeling constraints. This process produces a detailed task understanding for task $t$ and a fixed set of task-specific criteria, which are generated once and reused across all examples from the same task. Accordingly, each critic model input contains the task description, parent solution, proposed child solution, task understanding, and task-specific criteria, with example shown in Appendix ~\ref{app:task-understanding}.

\paragraph{Soft-rubric reasoning.}
The critic model follows a soft-rubric reasoning template. It first analyzes every criterion and compares the parent and child solutions under that criterion, then freely determines the relative importance of the criteria and aggregates its analyses into a final verdict. We call this design a \emph{soft rubric} because the critic model must evaluate the fixed criteria but remains unrestricted in its analysis and weighting strategy. Appendix~\ref{app:prompts} and Appendix~\ref{app:reasoning_examples} provide the complete prompt instructions and a reasoning example, respectively.

\subsection{Training Data and ML-Idea-Critic-Bench Construction}
\label{sec:data_collection}

\paragraph{Dataset overview.}
The dataset contains approximately 28K execution-labeled examples: approximately 12.5K in the SFT/evaluation pool and 15.5K in the GRPO training pool. We divide the former into approximately 11K SFT training examples and 1.5K static evaluation examples, which constitute \textsc{ML-Idea-Critic-Bench}. The evaluation set contains roughly 500 examples from each of the three teacher-assistance strata defined below.

\paragraph{ML idea-evolution trajectories.}
We collect parent--child pairs from MLE-Dojo~\citep{qiang2026mle} and MLS-Bench~\citep{lyu2026mls}, with Gemini-3.1-Pro generating both candidate ideas and their implementations. Initial parents are drawn from the baseline implementations provided by MLS-Bench or constructed as basic executable solutions for MLE-Dojo. Each trajectory example is represented as $x_i=(t_i,p_i,a_i,y_i)$, comprising a task description $t_i$, a parent idea $p_i$, a child idea $a_i$, and an outcome label $y_i$ derived from execution. An execution first yields one of three outcomes: \textsc{Improve}, \textsc{Tie}, or \textsc{Lose}. To provide more precise targets for critique synthesis in the SFT/evaluation pool, we use the empirical distribution of child performance relative to the parent to further divide \textsc{Improve} into \textsc{BigWin}, \textsc{MediumWin}, and \textsc{SmallWin}, yielding the five-way label space
\begin{equation*}
y_i \in
{\textsc{BigWin},\textsc{MediumWin},\textsc{SmallWin},
\textsc{Tie},\textsc{Lose}},
\end{equation*}
which captures both the direction and magnitude of the child's performance change relative to the parent. The GRPO pool retains the three-way outcome labels and is balanced across these classes by random downsampling. Detailed collection and label-construction are provided in Appendix~\ref{app:data}.

\paragraph{Teacher critique synthesis and difficulty strata.}
To collect high-quality SFT data, we use Gemini-3.1-Pro to generate soft-rubric reasoning critiques whose final predictions follow the same five-way label space. We first sample 8 critiques independently without revealing the execution-derived label $y_i$. An example is assigned to the \emph{Sampled} stratum if at least one of these critiques predicts $y_i$ correctly, in which case we retain one matching critique. If all 8 predictions are incorrect, we identify the most frequently predicted incorrect label, provide a hint explicitly ruling out that option, and sample another 8 critiques. An example is assigned to the \emph{Revised} stratum if at least one of these new critiques predicts $y_i$ correctly. If none succeeds, we reveal $y_i$ and require Gemini-3.1-Pro to generate reasoning consistent with the correct outcome; these examples form the \emph{Guided} stratum. The resulting SFT/evaluation pool contains three levels of increasing teacher assistance and example difficulty: \emph{Sampled} (easy, \textbf{44.4\%}), \emph{Revised} (hard, \textbf{31.6\%}), and \emph{Guided} (harder, \textbf{23.9\%}).

\subsection{Critic Model Training}

\paragraph{Supervised fine-tuning.}
We first fine-tune the critic model on approximately 11K Gemini-generated critiques collected using the procedure above, reserving the remaining 1.5K examples for static evaluation. By learning to produce both criterion-level analyses and final verdicts, the critic model learns the basic ability to evaluate ML proposals and follow the soft-rubric reasoning instructions.

\paragraph{Further Alignment with GRPO.}
Starting from the SFT checkpoint, we further train the critic model with Group Relative Policy Optimization (GRPO)~\citep{shao2024deepseekmath} on an additional 15.5K execution-labeled examples not used for SFT. We use a binary reward: a rollout receives $+1$ if its predicted three-way verdict matches the execution-derived label and $-1$ otherwise. Details of training are presented in Appendix~\ref{app:training_details}.

\section{ML-Idea-Critic Bench Result}
\label{sec:exp_setup}

\paragraph{Metrics.}
We report four complementary metrics. \textbf{Improve Accuracy} evaluates the binary decision of whether the child improves over its parent. \textbf{Dir-3} measures three-way accuracy over \emph{Improve}, \emph{Tie}, and \emph{Lose}. \textbf{F1} reports the F1 score for the binary improvement decision. Finally, \textbf{5-Ladder} evaluates the exact match accuracy over
[
\textsc{BigWin} >
\textsc{MediumWin} >
\textsc{SmallWin} >
\textsc{Tie} >
\textsc{Lose}
].

We compare our critics against Gemini-3.1-Pro, the base Qwen-3.5-9B and Qwen-3.5-35B models, and DeepSeek-GRM-16B/27B~\citep{liu2025inference}. DeepSeek-GRM is a generalist generative reward model trained with Self-Principled Critique Tuning to generate evaluation principles and critiques before assigning scores. We use the released DeepSeek-GRM-16B\footnote{\url{https://huggingface.co/BBQGOD/DeepSeek-GRM-16B}} and DeepSeek-GRM-27B\footnote{\url{https://huggingface.co/BBQGOD/DeepSeek-GRM-27B}} checkpoints, based on DeepSeek-V2-Lite and Gemma-2-27B, respectively. As shown in Table~\ref{tab:critic_main}, SFT improves the Dir-3 accuracy of the 9B and 35B critics to $55.6\%$ and $60.3\%$, surpassing Gemini-3.1-Pro ($46.6\%$). GRPO further improves their accuracy to $60.7\%$ and $63.3\%$, corresponding to gains of $5.1\%$ and $3.0\%$ over their SFT checkpoints and $14.1\%$ and $16.7\%$ over Gemini. 

\begin{table*}[t]
\centering
\caption{Idea-critic performance on the 1.5K-example evaluation set. Split-breakdown (Sampled/revised/guided) results are in Table~\ref{tab:difficulty_breakdown}. Best results are shown in bold.}
\label{tab:critic_main}
\begin{tabular}{llcccc}
\toprule
\textbf{Model} & \textbf{Training }
& \textbf{Acc}
& \textbf{Dir-3}
& \textbf{F1}
& \textbf{5-Ladder} \\
\midrule
Gemini-3.1-Pro      & Zero-shot                  & 54.3 & 46.6 & 59.8 & 28.2 \\
Qwen-3.5-9B         & Base                       & 49.9 & 41.3 & 58.3 & 22.4 \\
Qwen-3.5-35B        & Base                       & 49.9 & 36.4 & 51.8 & 20.1 \\
Deepseek-GRM-16B        & Ckpt from \citet{liu2025inference}     & 58.5 & 57.8 & 73.3 & 25.3 \\
Deepseek-GRM-27B        & Ckpt from \citet{liu2025inference}     & 58.4 & 58.1 & 73.6 & 25.3 \\
\midrule
Qwen-3.5-9B         & SFT             & 62.4 & 55.6 & 70.9 & 31.5 \\
Qwen-3.5-35B        & SFT             & 65.4 & 60.3 & 74.1 & 38.5 \\
\midrule
Qwen-3.5-9B         & SFT+GRPO        & 66.7 & 60.7 & \textbf{75.6} & 41.5 \\
Qwen-3.5-35B        & SFT+GRPO        & \textbf{68.5} & \textbf{63.3} & 75.4 & \textbf{45.4} \\
\bottomrule
\end{tabular}
\end{table*}

\label{sec:reasoning_ablation}
\paragraph{Ablation: Freestyle vs.\ Soft-Rubric Reasoning.} The two reasoning formats perform similarly after SFT, with Dir-3 accuracies of $54.6\%$ for freestyle reasoning and $55.6\%$ for soft-rubric reasoning. Under GRPO, however, the freestyle critic deteriorates to $40.3\%$, whereas the soft-rubric critic improves to $60.7\%$. Qualitatively, the freestyle critic develops shorter analyses and superficial shortcuts, while the required criterion-by-criterion comparisons stabilize reinforcement learning. Results with full metrics can be found in Appendix~\ref{app:additional_results}.

\section{Integration with Self-Evolving Agents}
\label{sec:openevolve_results}

\subsection{Inference-Time Idea Selection}\label{sec:openevolve-inference}

\paragraph{OpenEvolve setup.}
We integrate the critic model into OpenEvolve~\citep{sharma2025openevolve} to screen candidate ideas before empirical verification.  OpenEvolve is an evolutionary coding framework for recursive self-improvement that iteratively proposes, implements, and evaluates modifications as executable solutions. In this setup, Gemini-3.1-Pro serves as the idea proposer, implementer, and solution summarizer through three separate model calls. Each mutation follows the pipeline illustrated in Figure~\ref{fig:framework}: children are compared with their parent at the idea level, and child idea is empirically verified only after the screening from critic model:  The idea proposer Gemini-3.1-Pro is allowed to propose 16 ideas. For each idea, the critic model will make 16 reasoning critiques, and only the idea with the highest votes on win will be verified. For OpenEvolve setup, we maintain three islands and evolve each island for 10 rounds, yielding a total budget of 30 empirical verification runs per task. Additional details about the OpenEvolve configuration are provided in Appendix~\ref{app:openevolve-framework}. More details about critic model's idea-selection strategy are provided in Appendix~\ref{app:openevolve-idea-selection}.

\paragraph{Screening settings and evaluation tasks.}
We compare seven settings that differ only in how candidate ideas are screened. \emph{None} denotes vanilla OpenEvolve without idea-level screening. \emph{Gemini-3.1-Pro} and \emph{Gemini-3.5-Flash} use the corresponding pretrained models as critic models. \emph{Base-9B} and \emph{Base-35B} use the original Qwen-3.5-9B and Qwen-3.5-35B models, respectively, whereas \emph{Critic-9B} and \emph{Critic-35B} use our corresponding critic models trained with SFT and GRPO. We evaluate all settings on MLS-Lite-Bench, which contains 30 ML development tasks spanning 12 domains. Because every setting obtains a score of zero on \texttt{rl-value-discrete}, the benchmark's only reinforcement learning task, we exclude this non-discriminative task from aggregate reporting and present results over the remaining 29 tasks across 11 domains. The complete task configuration, domain grouping, and score normalization procedure are described in Appendix~\ref{app:mls-lite-details}.

Table~\ref{tab:openevolve-full} reports the MLS-Lite-Bench normalized score averaged within each domain. The settings using our trained critic models achieve the best score in 9 of the 11 domains and the highest overall average. This advantage persists on the 20 tasks excluded from critic model training: the 9B and 35B critic models achieve the two highest averages of 34.24 and 34.18, respectively, compared with 30.44 without screening. The largest gain appears in Language Models, where the normalized score increases from 18.04 without screening to 32.57 with the 9B critic model. Overall, the improvement margins are moderate because all settings share Gemini-3.1-Pro as the idea proposer and implementer. 


\begin{table}[t]
\centering
\newcolumntype{Y}{>{\centering\arraybackslash}X}
\caption{Openevolve with critic model integrated result with MLS-Lite-Bench with Gemini-3.1-Pro as idea proposer. All domains use 10 evolution rounds. The best result in each row is shown in \textbf{bold}.} \label{tab:openevolve-full} 
\setlength{\tabcolsep}{3pt} 
\renewcommand{\arraystretch}{1.08} 
\begin{tabularx}{\linewidth}{
    @{}
    >{\raggedright\arraybackslash}p{0.36\linewidth}
    *{7}{Y}
    @{}
}
\toprule 
\multirow{2}{*}{\textbf{Domain}} & \multirow{2}{*}{\textbf{None}} & \multicolumn{2}{c}{\textbf{Gemini}} & \multicolumn{2}{c}{\textbf{Qwen-3.5-9B}} & \multicolumn{2}{c}{\textbf{Qwen-3.5-35B}} \\ 
\cmidrule(lr){3-4} \cmidrule(lr){5-6} \cmidrule(l){7-8} & &
\mbox{\small\bfseries 3.1-Pro} &
\mbox{\small\bfseries 3.5-Flash} &
\mbox{\small\bfseries Base} &
\mbox{\small\bfseries Critic} &
\mbox{\small\bfseries Base} &
\mbox{\small\bfseries Critic} \\

\midrule 
Language Models                  & 18.04 & 18.85 & 17.66 & 32.48 & \textbf{32.57} & 32.54 & 31.47  \\ 
Robotics                         & 23.18 & 29.14 & 28.03 & 25.52 & \textbf{29.75} & 26.71 & 28.29 \\ 
Vision and Generation            & 16.11 & 15.04 & 15.08 & 15.48 & 17.38 & 14.86 & \textbf{17.84} \\ 
ML Systems and Efficient ML      & 64.00 & 64.01 & 63.73 & 63.11 & \textbf{65.54} & 63.59 & 63.80 \\ 
AI for Science                   & 33.55 & \textbf{34.52} & 32.38 & 33.94 & 31.67 & 32.68 & 33.09 \\ 
Optimization and Theory          & 22.46 & 24.69 & 24.85 & 22.23 & 24.79 & 24.64 & \textbf{24.86} \\ 
Classical and Adaptive Learning  & 51.85 & 51.26 & 51.50 & 45.25 & \textbf{53.53} & 45.43 & 47.40 \\ 
Time Series and Forecasting      & 54.64 & 52.73 & 51.54 & 52.14 & 57.24 & 48.18 & \textbf{58.72} \\ 
Structured and Causal Reasoning  & 55.81 & 54.35 & 55.93 & 52.83 & 54.01 & 52.61 & \textbf{56.44} \\ 
Deep Learning                    & 60.01 & \textbf{64.90} & 63.01 & 62.49 & 62.67 & 60.94 & 62.27 \\
Trustworthy Learning             & 26.85 & 28.84 & 30.32 & 25.91 & 27.46 & 29.60 & \textbf{30.99} \\
\midrule 
Domain Average                   & 38.77 & 39.85 & 39.46 & 39.22 & \textbf{41.51} & 39.25 & 41.38 \\ 
Heldout-Only Average             & 30.44 & 32.12 & 31.15 & 32.51 & 34.24 & 32.68 & 34.18 \\
\bottomrule 
\end{tabularx} 
\end{table}
\subsection{Test Time Continual Learning}

\begin{table}[h] \centering  \caption{Task-specific test-time continual learning result. TT stands for test time continual learning alignment.  Scores higher than no continual learning baseline are shown in \textbf{bold}.} \label{tab:test-time} 
\renewcommand{\arraystretch}{1.08} 
\begin{tabularx}{\linewidth}{ @{} >{\raggedright\arraybackslash}X cccccc @{} } 
\toprule 
\multirow{2}{*}{\textbf{Task}} & \multicolumn{2}{c}{\textbf{Frozen Critic}} & \multicolumn{2}{c}{\textbf{TT-GRPO}} & \multicolumn{2}{c}{\textbf{TT-SDFT}} \\ 
\cmidrule(lr){2-3} \cmidrule(lr){4-5} \cmidrule(l){6-7} & \textbf{9B} & \textbf{35B} & \textbf{9B} & \textbf{35B} & \textbf{9B} & \textbf{35B} \\ 
\midrule 
ML Dimensionality Reduction        & 49.62 & 38.43 & \textbf{50.06} & \textbf{42.15} & \textbf{50.43} & \textbf{51.77} \\ 
TS Exogenous Forecast              & 48.43 & 49.57 & \textbf{48.90} & 47.02 & \textbf{49.11} & \textbf{51.17} \\ 
AI4Bio Mutation Effect Prediction  & 48.01 & 49.88 & 47.34 & 49.47 & \textbf{51.66} & \textbf{50.34}\\ 
Graph Generation                   & 70.69 & 73.30 & \textbf{72.58} & 73.05 & \textbf{73.51} & \textbf{76.64} \\ \bottomrule \end{tabularx} \end{table}

We investigate whether adapting an idea-level critic during agent evolution improves the final solution quality under a fixed verification budget. We consider two strategies: test-time Group Relative Policy Optimization (GRPO)~\citep{yuksekgonul2026learning} and Self-Distillation Fine-Tuning (SDFT)~\citep{shenfeld2026self}. We pick 4 tasks from MLS-Lite-Bench to evaluate both strategies, under the same OpenEvolve setup in Section~\ref{sec:openevolve-inference}. Both methods start from our trained critic model checkpoint, keep its backbone parameters frozen, and optimize a LoRA adapter with rank 32. The updates use only outcomes obtained from the empirical verifications already performed by OpenEvolve and therefore require no additional verifier calls.

\paragraph{Self-Distillation Fine-Tuning.}
Self-Distillation Fine-Tuning (SDFT) is introduced by \citet{shenfeld2026self} for continual learning from demonstrations. SDFT uses the same underlying model as both student and teacher, but provides the teacher with privileged context and distills its token-level predictions into the student along trajectories generated by the student itself. We adapt this idea to test-time alignment by treating the raw scores of parents and verified children as privileged teacher information. 
At every prefix of the student-generated critique, the teacher produces a next-token distribution, which is the fixed target for updating the student. We minimize the resulting token-level reverse-KL loss over the critique trajectory. We perform one SDFT update after every OpenEvolve round using the three newly verified parent--child pairs, one from each island.

\paragraph{Test-time GRPO.}
TTT-Discover~\citep{yuksekgonul2026learning} formulated scientific discovery as test-time RL, in which a policy model updates its parameters using experience collected while solving the problem. Following this paradigm, we apply GRPO to the critic model during OpenEvolve. We perform one update every 2 evolution rounds, collecting 6 newly verified parent--child pairs across the three islands. For each pair, we sample 16 critic model rollouts and assign each rollout a reward of +1 if its predicted three-way outcome matches the empirical label and -1 otherwise. 

As shown in Table~\ref{tab:test-time}, SDFT improves downstream evolution more consistently than test-time GRPO. One possible explanation is the limited information in binary correctness rewards combined with the changing outcome distribution during evolution. When all 16 GRPO rollouts receive the same reward, their group-relative advantages vanish, providing no learning signal from the correctness reward. As the parent solution improves, successful modifications can also become rarer, encouraging the critic model to rely on the increasing prevalence of \textsc{Tie/Lose} rather than distinguish promising candidates. Such behavior may achieve high prediction accuracy while providing little value for candidate selection. SDFT instead supplies outcome-conditioned token-level targets along the student's reasoning trajectory, offering a richer adaptation signal than verdict correctness alone. Our results align with existing work~\citep{shenfeld2026self} that this form of supervision can better support candidate selection and improve final solution quality under a limited verification budget.

\section{The Critic Serving as Reward Model}

While critic models help OpenEvolve select more promising ideas, its performance remains constrained by the proposal capabilities of the fixed policy model. We therefore investigate whether critic model predictions can provide useful training signals to improve the policy model's ability to generate better ideas.

\subsection{GRPO with Critic-in-Loop}

We further investigate whether the trained critic model can serve as a learned reward model for aligning the idea-proposing policy. We initialize the policy from the base Qwen-3.5-9B and  Qwen-3.5-35B model and optimize it using GRPO~\citep{shao2024deepseekmath} across the same 95 MLE-Dojo and 46 MLS-Bench training environments used for critic GRPO training.  Each training step collects 16 parents, each with 16 rollout ideas, yielding 256 candidate ideas per step. The reward is binary: a solution receives $+1$ for an \textsc{Improve} outcome and $-1$ otherwise. We consider two approaches, (1) baseline approach \emph{verify-all}, where every candidate is empirically evaluated; and (2) \emph{critic-in-loop} approach, where trained critic models serve as a reward model and replace real verification signals occasionally, depending on the critic's uncertainty on the candidate idea. Specifically, we use Critic-9B checkpoint serving base Qwen-3.5-9B and Critic-35B checkpoint serving base Qwen-3.5-35B respectfully. The critic models keep their parameters frozen throughout policy training. 
For each candidate $a$, let $v_y(a)=\frac{1}{16}\sum_{k=1}^{16}\mathbb{I}[\tilde{y}_k(a)=y]$ denote the fraction of critic model rollouts $\tilde{y}_k(a)$ predicting label $y$. We define uncertainty as $u(a)=1-\max_y v_y(a)$, namely, the fraction of rollouts that disagree with the majority. If at most three of the 16 rollouts disagree ($u(a) < 0.25$), we use the majority verdict to calculate reward; otherwise, we send the candidate to the ground-truth verifier.

\subsection{Policy Model GRPO Result}

For a controlled comparison, each method receives a total budget of 10,000 ground-truth verifications across all training environments. With 256 candidates per training step, \emph{verify-all} exhausts this budget in approximately 40 steps. With critic-guided reward assignment, only $36.3\%$ of policy-generated ideas require empirical verification, allowing approximately 110 steps under the same budget.  This corresponds to a $2.75\times$ improvement in verification efficiency, measured as the number of candidate ideas assigned training rewards per ground-truth verification. This ratio does not measure wall-clock speedup or total compute savings, since critic inference incurs additional computation cost.

\begin{figure}[t]
    \centering
    \begin{minipage}[t]{0.58\textwidth}
        \vspace{0pt}
        \centering
        \includegraphics[width=\linewidth]{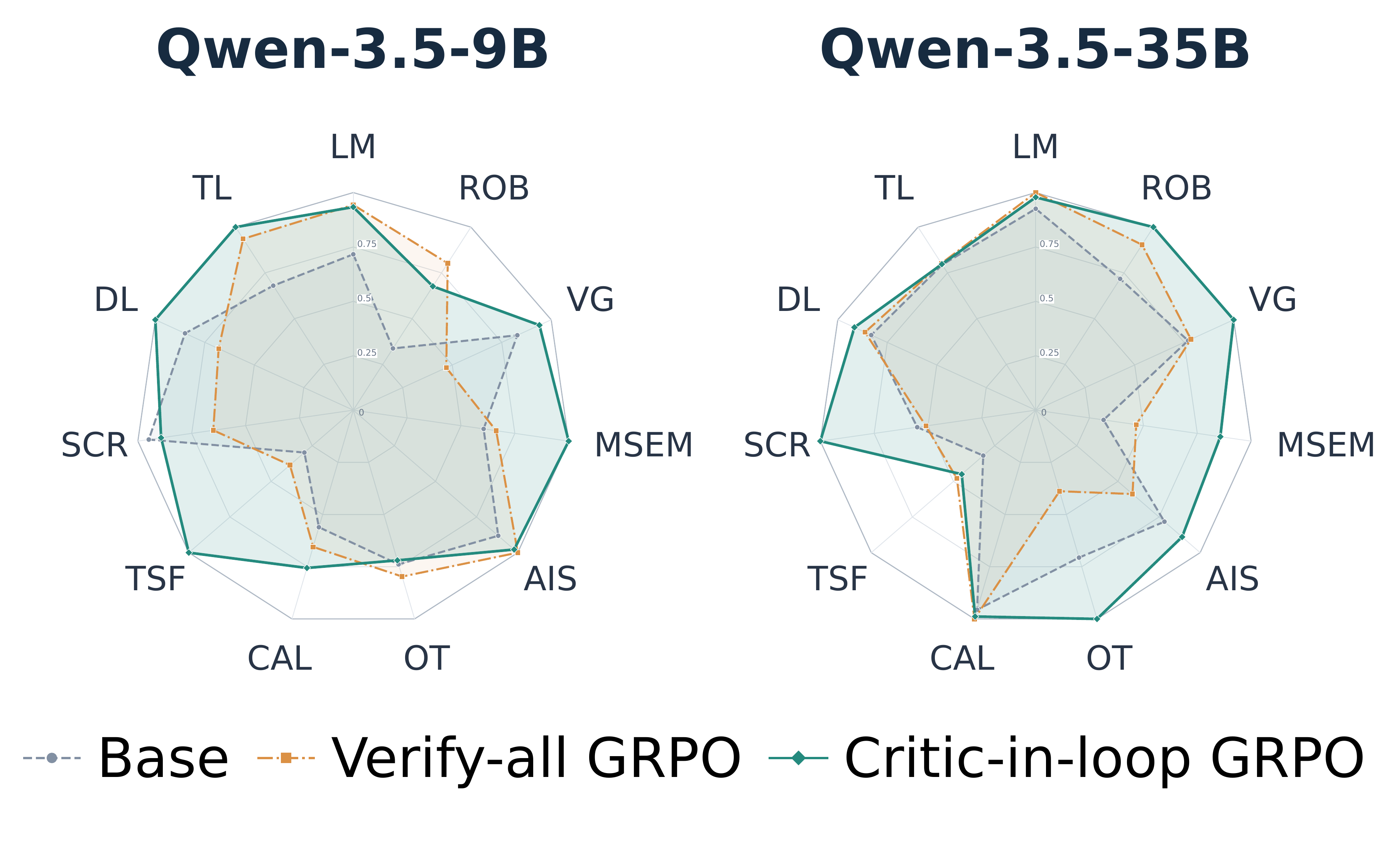}
    \end{minipage}\hfill%
    \begin{minipage}[t]{0.40\textwidth}
        \vspace{0pt}
        \centering
        \begingroup
\centering 
\small 
\label{tab:policy-grpo} 
\setlength{\tabcolsep}{3pt} 
\begin{tabular}{l l} 
\toprule 
\textbf{Code} & \textbf{Domain} \\

\midrule 
LM & Language Models                  \\ 
ROB & Robotics                         \\ 
VG & Vision and Generation             \\ 
MSEM & ML Systems and Efficient ML      \\ 
AIS & AI for Science                   \\ 
OT & Optimization and Theory          \\ 
CAL & Classical and Adaptive Learning  \\ 
TSF & Time Series and Forecasting      \\ 
SCR & Structured and Causal Reasoning  \\ 
DL & Deep Learning                     \\
TL & Trustworthy Learning            \\
\bottomrule 
\end{tabular} 
\endgroup
    \end{minipage}
    \vspace{-10pt}
    \caption{Trained idea-proposing policies performance on MLS-Lite-Bench. For Qwen-3.5-9B (left) and Qwen-3.5-35B (right), we compare the base policy, verify-all GRPO, and critic-model-in-the-loop GRPO across 11 domains. For each domain, scores are normalized against the best of the six arms.}
        \label{fig:policy-grpo}
    \vspace{-5pt}
\end{figure}

We evaluate the resulting policies directly on the MLS-Lite-Bench tasks. As shown in Figure~\ref{fig:policy-grpo}, critic-in-loop GRPO improves the average normalized domain score from 19.57 to 25.21 for the 9B model and from 19.29 to 24.55 for the 35B model compared with verify-all GRPO. These results suggest that using critic-model predictions for confident cases allows the saved verification budget to support additional policy updates and thereby shifts the policy toward more promising ML modifications. Full numerical results are reported in Table~\ref{tab:policy-grpo-table} of Appendix~\ref{app:policy-grpo-table}.

\subsection{Ablation Experiments}

\paragraph{Uncertainty Analysis. }
We analyze how the number of critic rollouts $k$ and the uncertainty threshold $T$ affect the reliability of majority-vote prediction. 
We evaluate how $k$ and $T$ influencing the following three quantities: the AUROC of the vote-based critic prediction, the fraction of critic errors recalled by uncertainty-based routing, and the fraction of all ideas sent to the verifier.

\begin{figure}[h]
    \centering
    \includegraphics[width=\linewidth]{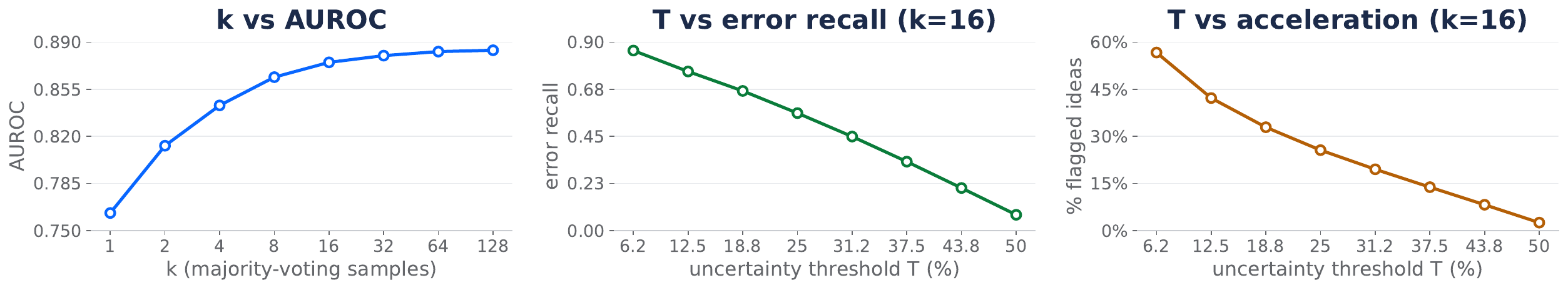}
    \vspace{-15pt}
    \caption{Uncertainty analysis. Increasing the number of critic samples yields diminishing AUROC gains beyond $k=16$ (left). At $k=16$, an uncertainty threshold of $0.25$ recalls around 70\% of incorrect predictions while routing about 30\% of all ideas to empirical verification (middle and right). }
    \label{fig:critic-uncertainty}
\end{figure}

Figure~\ref{fig:critic-uncertainty} reports the sweep on the test split of \textsc{ML-Idea-Critic-Bench}. Increasing $k$ initially produces substantial improvements in the AUROC of the vote-based critic prediction, but the gains largely saturate at $k=16$.
With an uncertainty threshold around $T=0.25$, the selective-verification rule identifies approximately 70\% of the critic's errors while routing only about 30\% of all ideas to the ground-truth verifier.  This corresponds to our critic-in-loop experiments, 36.3\% of policy-generated ideas are empirically verified, moderately exceeding the 30\% estimated from the benchmark. 

\paragraph{Ablation on the Critic as a Reward Model.}
We test whether critic model predictions alone can drive policy optimization. Starting from the base Qwen-3.5-35B policy, we perform GRPO using rewards provided by the frozen Critic-35B without real verification. We conduct two sweeps: varying the number of policy training steps from 0 to 160, and varying $k\in\{1,4,8,16,32\}$ while fixing training to 40 steps, where $k$ is the number of critique rollouts for majority voting. Figure~\ref{fig:critic-reward-ablation} in Appendix~\ref{app:ablation-critic} shows that policy performance generally improves as the number of training steps increases on two selected tasks. Increasing the number of critic model rollouts $k$ also improves performance on the selected tasks, but the gains are comparatively modest and begin to saturate. In particular, increasing $k$ beyond 16 provides limited additional benefit while proportionally increasing inference computational cost. 
\vspace{-5pt}

\section{Related Work}
\paragraph{Self-Evolving Agents and ML Agents.}
ML and self-evolving agents iteratively propose, implement, and evaluate solutions using empirical feedback~\citep{huang2024mlagentbench,jiang2025aide,novikov2025alphaevolve,sharma2025openevolve}.
Existing approaches improve this process through search and harness design with fixed pretrained models~\citep{toledo2026ai,yan2026pacevolve,yamada2025ai}, or by learning proposal policies from feedback~\citep{zhang2026learning,yan2026pacevolve++}.
Our work complements these directions by learning to evaluate proposed modifications before costly implementation and execution.

\paragraph{AI for Machine Learning Research.}
AI-assisted research systems coordinate literature review, experimentation, and manuscript preparation~\citep{schmidgall2025agent,tang2026ai}, while related work studies automated paper reviewing~\citep{tang2026fars,liu2026autoresearchclaw} and research-idea evaluation~\citep{qiao2026innoeval,moussa2025scholareval}.
Closer to our setting, pretrained language model-based evaluation modules assess ML candidates and prioritize promising ones for execution~\citep{foster2026ai,zheng2026can}.
We instead make the evaluator itself the primary learned component, training an idea-level critic model to predict the empirical value of proposed ML modifications.

\paragraph{Critic Models and Generative Reward Models.}
Conventional reward models typically assign scalar scores to model outputs~\citep{ouyang2022training,lambert2025rewardbench}.
Critic models and generative reward models produce textual assessments that make the basis of their judgments more explicit~\citep{saunders2022self,mcaleese2024llm,zhang2025generative,chen2026rm}.
These assessments support direct output evaluation~\citep{zheng2023judging,ke2024critiquellm}, actionable feedback for iterative refinement~\citep{wang2023shepherd,xie2025teaching,yang2025lighthouse}, and ranking among sampled candidates~\citep{liu2025inference,zhang2025generative,saha2025learning}.
Rubric-guided approaches further structure evaluations across multiple criteria~\citep{viswanathan2025checklists,gunjal2026rubrics,guan2026evorubric,qiu2026rationale}.
Related work evaluates intermediate reasoning steps~\citep{uesato2022solving,lightman2024let} or uses learned evaluations as reward signals for policy optimization and co-evolution~\citep{shan2026bringing,li2026no,chen2025spc,zhang2026qevolve}.
We focus on proposed modifications to existing ML solutions, whose ground-truth value requires empirical execution.
Our critic model learns from empirical parent--child outcomes, supporting both inference-time candidate selection and policy training.

More discussion on related work is in Appendix~\ref{app:more-related-work}.

\section{Conclusion}
We introduced idea-level critic models as an efficient feedback mechanism for self-improving ML agents together with a training pipeline combining soft-rubric reasoning, SFT, and GRPO. Our critic model not only achieve strong performance in static idea evaluation, but also improve agents across inference-time idea selection, task-specific continual learning, and critic-guided policy optimization. By identifying promising ideas, the critic model enables agents to use limited verification budgets more effectively. As a learned reward model, the critic model also performs \(2.75\times\) more policy-optimization steps under the same budget. Overall, our results establish idea-level critic model as a general and practical interface between idea generation and expensive execution, enabling more efficient ML agent learning and evolution.

\clearpage
\newpage
\bibliographystyle{assets/plainnat}
\bibliography{paper}

\clearpage
\newpage
\beginappendix

\section{Appendix}
\appendix

\section{More Related Work}\label{app:more-related-work}

\paragraph{Self-Evolving Agents and ML Agents.}
ML agents automate ML development by translating task specifications into executable solutions, running training and evaluation pipelines, interpreting empirical results, and iteratively revising their approaches~\citep{huang2024mlagentbench,chan2025mlebench,qiang2026mle,jiang2025aide,qiao2026scaling,yu2025alpharesearch,yu2026mltoolbenchlearningtoolaugmentedagents}. Closely related self-evolving and recursively self-improving agents repeatedly modify solutions using feedback from previous iterations~\citep{lu2024aiscientist,zhang2025darwingodel,novikov2025alphaevolve,sharma2025openevolve}. These systems commonly follow a propose--implement--execute--refine loop, and recent work increasingly targets the generation and selection of promising candidates before execution. Some approaches improve the surrounding agent harness while keeping the pretrained model fixed, using structured search and refinement to explore stronger solutions~\citep{toledo2026ai,nam2026mle,yan2026pacevolve,yamada2025ai}; others learn proposal policies directly from feedback~\citep{zhang2026learning,yan2026pacevolve++}.

\paragraph{AI for Machine Learning Research.}
LLM-based agents increasingly automate machine learning research, from literature review and ideation to experimentation, analysis, and manuscript preparation~\citep{lu2024aiscientist, schmidgall2025agent, tang2026ai}.
Related systems connect idea generation to execution, introduce falsification, and support collaborative or sustained research~\citep{li2024mlr, liu2024aigs, schmidgall2025agentrxiv, weng2026deepscientist}.
Recent work further explores research automation at scale and iterative improvement through human feedback and accumulated experience~\citep{tang2026fars, liu2026autoresearchclaw}.
A complementary line develops AI systems for paper reviewing through multi-agent feedback, aspect-conditioned generation, structured questioning, and rubric-guided evaluation~\citep{d2024marg, gao2024reviewer2, gao2025reviewagents, chang2025treereview, li2026reviewgrounder}.
Learned reviewing models also support iterative refinement of research outputs~\citep{weng2025cycleresearcher}, while empirical studies examine the effects of LLM use in peer review~\citep{kim2026use}.
Closer to our setting, pretrained language model-based evaluation modules assess unverified ideas or candidate ML solutions to prioritize empirical execution~\citep{weng2026deepscientist, zheng2026can, foster2026ai}.
Our work focuses on training an execution-grounded idea-level critic model to predict the empirical value of proposed ML modifications.
The critic model evaluates each modification relative to an existing solution and learns from outcome labels derived from empirical parent--child comparisons.
We develop a training recipe for this prediction task and apply the resulting critic model to candidate selection during inference-time evolution and reward provision during policy optimization.

\paragraph{{Critic Models and General Reward Models.}} 

Conventional reward models compress the quality of a model output into a scalar score, providing limited insight and often struggling with open-ended problems whose success depends on multiple interacting factors~\citep{ouyang2022training, lambert2025rewardbench}. 
Critic models and general reward models instead generate reasoning-style analyses before reaching a final judgment~\citep{yao2022react}, making their evaluations more interpretable and allowing them to produce more reliable, task-general feedback~\citep{liu2025inference,saunders2022self, zheng2023judging, mcaleese2024llm, zhang2025generative, chen2026rm}.  Existing work uses such generative evaluators in several related ways. 
At inference time, they can directly assess completed outputs~\citep{zheng2023judging, ke2024critiquellm}, provide actionable feedback for iterative refinement~\citep{saunders2022self, mcaleese2024llm, wang2023shepherd, xie2025teaching, yang2025lighthouse,jin2026weak}, or rank multiple sampled candidates so that only the most promising response is returned~\citep{liu2025inference, zhang2025generative, saha2025learning}. 
In structured scenarios, rubric-guided critiques further regularize multi-dimensional evaluations and mitigate evaluation degradation~\citep{liu2025inference, viswanathan2025checklists, gunjal2026rubrics, wang2026co, guan2026evorubric,wang2026serpo, qiu2026rationale, chen2026graph,yu2026reinforcement}. 
Other approaches formulate generative critics more explicitly as value models: they estimate the quality or expected return of intermediate states and trajectories~\citep{uesato2022solving, lightman2024let}, and their judgments can be converted into learning signals for policy optimization, including PPO-style training and adversarial co-evolution~\citep{shan2026bringing,wang2026co, li2026no, chen2025spc,zhang2026qevolve}. 
These methods have primarily been developed to evaluate language or reasoning outputs produced by policy models. 
A separate line of work evaluates research idea~\citep{qiao2026innoeval, moussa2025scholareval, falck2026training, liu2026guide,wang2026ligbench} by constructing methodology on human level ideas, which are more tailored to AutoResearch and human-AI collaboration research, which is beyond the scope of our paper.

\section{Data Collection and Label Construction}
\label{app:data}

\subsection{Task Environments and Initial Parent Solutions}

We construct our data from selected task environments in MLE-Dojo~\citep{qiang2026mle} and MLS-Bench~\citep{lyu2026mls}. MLS-Bench provides source code and baseline implementations for each task, which we directly use as the initial parent solutions. MLE-Dojo provides executable task environments but does not include comparable starter solutions. For each selected MLE-Dojo task, we therefore prompt Gemini-3.1-Pro to generate a basic runnable solution, execute it in the corresponding environment, and use its resulting score as the initial parent score.

\subsection{Trajectory Collection and Label Construction}
\label{app:collection_labeling}

\paragraph{SFT/evaluation pool: two-stage collection.}
We construct the SFT/evaluation pool using a two-stage procedure. In the first stage, we run ten sequential evolution steps on each selected task, starting from its initial executable solution. The solution produced at each step is retained as a parent for the second stage. In the second stage, we independently propose and execute 20 child solutions for each retained parent, with all 20 children evaluated against the same parent. Gemini-3.1-Pro generates the ideas and implements the solutions
in both stages. The resulting parent--child pairs form the candidate pool for SFT and static evaluation.

\paragraph{Parent-specific labels.}
For each parent $p$, we compute the signed relative improvements of its executed children, orienting the task metric so that positive values indicate better performance than the parent. We exclude infeasible child implementations and parents with fewer than three improving children. The child outcomes associated with each retained parent provide the empirical mean $\mu_p$ and standard deviation $\sigma_p$ used to construct the five-level labels.

\paragraph{GRPO pool: improvement-driven collection.}
We collect GRPO training examples using a separate iterative procedure across the designated training tasks. At each step, Gemini-3.1-Pro proposes and implements one child solution from the current parent. If empirical evaluation shows an improvement, the child becomes the parent for the next step; otherwise, the current parent is retained. This procedure collects examples across a changing set of parents as successful modifications accumulate.

The initial GRPO pool contains approximately 25K execution-labeled examples, with an imbalanced distribution of \textsc{Improve}, \textsc{Tie}, and \textsc{Lose} outcomes. We randomly downsample within outcome classes to obtain a balanced training set of 15,669 examples, comprising 5,223 examples per class, as imbalanced distribution will cause critic model GRPO training to always output the majority label. This pool is used for critic GRPO training and is separate from the SFT/evaluation pool.

\subsection{Parent-Specific Label Thresholds, Dataset Statistics, and Label Distribution}

The same absolute improvement can have different significance depending on the quality and difficulty of the current parent solution. We therefore construct parent-specific labels. For each parent $p$, we generate and execute 20 child solutions and compute their signed relative improvements $\Delta_{p,1},\ldots,\Delta_{p,20}$, orienting each task metric so that a positive value always indicates improvement. We filter out parents that have less than 3 improved children solutions. We approximate the resulting improvement distribution for each parent with a Gaussian whose empirical mean and standard deviation are denoted by $\mu_p$ and $\sigma_p$. The standardized improvement of child $j$ is
\[
z_{p,j}=\frac{\Delta_{p,j}-\mu_p}{\sigma_p}.
\]

We assign the five-level label in descending order, using the first condition that applies:
\[
y_{p,j} =
\begin{cases}
\textsc{BigWin},
& \Delta_{p,j}>0
  \ \text{and}\ 
  \left(\Delta_{p,j}>0.25 \ \text{or}\ z_{p,j}>2\right),\\
\textsc{MediumWin},
& \Delta_{p,j}>0
  \ \text{and}\ 
  \left(\Delta_{p,j}>0.10 \ \text{or}\ z_{p,j}>1\right),\\
\textsc{SmallWin},
& \Delta_{p,j}>0
  \ \text{and}\ z_{p,j}>0.2,\\
\textsc{Tie},
& \Delta_{p,j}\geq 0 \ \text{or} \ -0.2\leq z_{p,j}\leq 0.2,\\
\textsc{Lose},
& \text{otherwise}.
\end{cases}
\]
The absolute relative-improvement thresholds identify large gains even for parents whose child scores have high variance, while the standardized thresholds adapt the labels to the local difficulty of improving each parent. The Gaussian approximation is used only to compute the parent-specific standardized score; the final labels are assigned deterministically using the rules above.

After removing infeasible child solutions, the resulting SFT dataset. The five labels in SFT dataset account for \textbf{25.4\%} \textsc{BigWin}, \textbf{17.5\%} \textsc{MediumWin}, \textbf{15.5\%} \textsc{SmallWin}, \textbf{13.6\%} \textsc{Tie}, and \textbf{28.0\%} \textsc{Lose}.

\section{Templates and Outputs Examples}

\subsection{ML Agent Pipeline Examples}\label{app:pipeline-example}

We present a recorded example of the ML agent pipeline on the \texttt{dl-activation-function} task from MLS-Lite-Bench. The example follows one candidate from idea proposal through implementation, summarization, critic model screening, and empirical verification. The tables retain the recorded prompts and generated outputs, with repeated inputs referenced across tables.

\paragraph{Task and parent solution.}
Table~\ref{tab:pipeline-task-parent} presents the task description and the initial parent program. The task is to design a custom activation function while keeping the model architectures, optimizer, and data pipeline fixed. The parent uses ReLU and achieves a \texttt{combined\_score} of $0.5997$.

\paragraph{Idea proposal.}
Table~\ref{tab:pipeline-proposal} shows the  instructions given to the idea proposer and its generated approach. Conditioned on the task and parent program, the proposer suggests Mish-ACON-C, a Mish-based activation with bounded learnable parameters that control its slopes and transition sharpness.

\paragraph{Implementation.}
Table~\ref{tab:pipeline-implementation} shows how the committed idea is translated into executable code. The implementer receives the proposed approach and the parent program, and replaces only the editable \texttt{CustomActivation} class. This implementation is produced before critic model screening and is reused if the candidate is selected for empirical verification.

\paragraph{Solution summarization.}
Table~\ref{tab:pipeline-summarization} presents the summarization instructions and the resulting child description. The summarizer receives both the proposed approach and its implementation, and is instructed to describe what the code does without predicting its performance. The resulting code-grounded summary serves as the child input to the critic model.


\newpage
\begingroup
\footnotesize
\setlength{\LTleft}{0pt}
\setlength{\LTright}{0pt}
\setlength{\LTcapwidth}{\linewidth}
\setlength{\LTpre}{6pt}
\setlength{\LTpost}{10pt}
\renewcommand{\arraystretch}{1.08}
\begin{longtable}{@{}>{\raggedright\arraybackslash}p{\linewidth}@{}}
\caption{Running example: task context and parent program for task activation-function design. }\label{tab:pipeline-task-parent}\\
\toprule
\textbf{Task and parent program}\tabularnewline
\endfirsthead
\multicolumn{1}{@{}l@{}}{\footnotesize\tablename~\thetable{} (continued)}\\
\toprule
\textbf{Task and parent program(DL Activation Function Design)}\tabularnewline
\midrule
\endhead
\midrule
\multicolumn{1}{@{}r@{}}{\footnotesize\emph{Continued on the next page.}}\\
\endfoot
\bottomrule
\endlastfoot
\midrule
\textbf{Research Question}\tabularnewline*[2pt]
Design an activation function for deep convolutional neural networks that improves test accuracy across different architectures (ResNet, VGG) and datasets (CIFAR-10, CIFAR-100, FashionMNIST), while keeping the model definitions, optimizer, initialization, and data pipeline fixed.\tabularnewline[3pt]
\textbf{Background}\tabularnewline*[2pt]
Activation functions introduce nonlinearity into neural networks and critically affect training dynamics, gradient flow, sparsity, and generalization. Classic and modern choices include:\tabularnewline[3pt]
- \textbf{ReLU}: \lstinline[basicstyle=\ttfamily\footnotesize,columns=fullflexible,breaklines=true,breakatwhitespace=false]|max(0, x)| \textemdash{} simple, sparse, but zero gradient for negative inputs .\tabularnewline[2pt]
- \textbf{GELU}: \lstinline[basicstyle=\ttfamily\footnotesize,columns=fullflexible,breaklines=true,breakatwhitespace=false]|x * Phi(x)| where \lstinline[basicstyle=\ttfamily\footnotesize,columns=fullflexible,breaklines=true,breakatwhitespace=false]|Phi| is the standard Gaussian CDF.\tabularnewline[2pt]
- \textbf{Swish / SiLU}: \lstinline[basicstyle=\ttfamily\footnotesize,columns=fullflexible,breaklines=true,breakatwhitespace=false]|x * sigmoid(beta * x)|; self-gated, smooth, non-monotonic.\tabularnewline[2pt]
- \textbf{Mish}: \lstinline[basicstyle=\ttfamily\footnotesize,columns=fullflexible,breaklines=true,breakatwhitespace=false]|x * tanh(softplus(x))|; self-regularized, smooth, non-monotonic.\tabularnewline[2pt]
- \textbf{Squared ReLU}, \textbf{StarReLU}, and other variants explore polynomial gates and learnable/affine extensions.\tabularnewline[2pt]
\textbf{What You Can Modify}\tabularnewline*[2pt]
The \lstinline[basicstyle=\ttfamily\footnotesize,columns=fullflexible,breaklines=true,breakatwhitespace=false]|CustomActivation| class inside \lstinline[basicstyle=\ttfamily\footnotesize,columns=fullflexible,breaklines=true,breakatwhitespace=false]|pytorch-vision/custom_activation.py|. It is an \lstinline[basicstyle=\ttfamily\footnotesize,columns=fullflexible,breaklines=true,breakatwhitespace=false]|nn.Module| used as a drop-in replacement for ReLU throughout the network.\tabularnewline[3pt]
You may modify the \lstinline[basicstyle=\ttfamily\footnotesize,columns=fullflexible,breaklines=true,breakatwhitespace=false]|forward| computation (any element-wise or channel-wise operation), register learnable parameters in \lstinline[basicstyle=\ttfamily\footnotesize,columns=fullflexible,breaklines=true,breakatwhitespace=false]|__init__|, choose any shape of activation curve (monotonic / non-monotonic / bounded), and decide negative-domain behavior (zero, linear, bounded, learnable). Tensor shape must be preserved.\tabularnewline[3pt]
The activation is used in:\tabularnewline[3pt]
- ResNet: BasicBlock (twice per block) and the initial conv.\tabularnewline[2pt]
- VGG: after every Conv-BN pair and inside the classifier head.\tabularnewline[2pt]
- MobileNetV2: replaces the ReLU6 baseline used in inverted residuals.\tabularnewline[2pt]
\textbf{Fixed Pipeline}\tabularnewline*[2pt]
- Optimizer: SGD with \lstinline[basicstyle=\ttfamily\footnotesize,columns=fullflexible,breaklines=true,breakatwhitespace=false]|lr=0.1|, \lstinline[basicstyle=\ttfamily\footnotesize,columns=fullflexible,breaklines=true,breakatwhitespace=false]|momentum=0.9|, \lstinline[basicstyle=\ttfamily\footnotesize,columns=fullflexible,breaklines=true,breakatwhitespace=false]|weight_decay=5e-4|.\tabularnewline[2pt]
- Schedule: cosine annealing over \lstinline[basicstyle=\ttfamily\footnotesize,columns=fullflexible,breaklines=true,breakatwhitespace=false]|200| epochs.\tabularnewline[2pt]
- Data augmentation: \lstinline[basicstyle=\ttfamily\footnotesize,columns=fullflexible,breaklines=true,breakatwhitespace=false]|RandomCrop(32, pad=4)| + \lstinline[basicstyle=\ttfamily\footnotesize,columns=fullflexible,breaklines=true,breakatwhitespace=false]|RandomHorizontalFlip|.\tabularnewline[2pt]
- Weight initialization: standard Kaiming normal (fixed).\tabularnewline[2pt]
- Evaluation settings: ResNet-20 on CIFAR-10, VGG-16-BN on CIFAR-100, MobileNetV2 on FashionMNIST.\tabularnewline[2pt]
\textbf{Baselines}\tabularnewline*[2pt]
- \textbf{gelu} \textemdash{} Hendrycks \& Gimpel, arXiv:1606.08415; \lstinline[basicstyle=\ttfamily\footnotesize,columns=fullflexible,breaklines=true,breakatwhitespace=false]|nn.GELU| (no learnable parameters).\tabularnewline[2pt]
- \textbf{silu} \textemdash{} Ramachandran et al. / Elfwing et al., arXiv:1710.05941; \lstinline[basicstyle=\ttfamily\footnotesize,columns=fullflexible,breaklines=true,breakatwhitespace=false]|nn.SiLU|, equivalent to Swish with \lstinline[basicstyle=\ttfamily\footnotesize,columns=fullflexible,breaklines=true,breakatwhitespace=false]|beta=1| (no learnable parameters).\tabularnewline[2pt]
- \textbf{mish} \textemdash{} Misra, arXiv:1908.08681; \lstinline[basicstyle=\ttfamily\footnotesize,columns=fullflexible,breaklines=true,breakatwhitespace=false]|x * tanh(softplus(x))| (no learnable parameters).\tabularnewline[2pt]
\textbf{Metric}\tabularnewline*[2pt]
Best test accuracy (\%, higher is better) achieved during training. The activation must be differentiable, shape-preserving, and must not change normalization layers, residual blocks, classifier heads, datasets, or the training loop.\tabularnewline[3pt]
\midrule
\textbf{Parent editable band (source Section 0.2)}\tabularnewline*[3pt]
\lstinline[basicstyle=\ttfamily\footnotesize,columns=fullflexible,keepspaces=true,showstringspaces=false,breaklines=true,breakatwhitespace=false]|class CustomActivation(nn.Module):|\tabularnewline[-2pt]
\hspace*{2.0em}\lstinline[basicstyle=\ttfamily\footnotesize,columns=fullflexible,keepspaces=true,showstringspaces=false,breaklines=true,breakatwhitespace=false]|"""Custom activation function. Drop-in replacement for ReLU.|\tabularnewline[-2pt]
\hspace*{2.0em}\lstinline[basicstyle=\ttfamily\footnotesize,columns=fullflexible,keepspaces=true,showstringspaces=false,breaklines=true,breakatwhitespace=false]|Input/Output: tensor of any shape, element-wise operation.|\tabularnewline[-2pt]
\hspace*{2.0em}\lstinline[basicstyle=\ttfamily\footnotesize,columns=fullflexible,keepspaces=true,showstringspaces=false,breaklines=true,breakatwhitespace=false]|Design considerations:|\tabularnewline[-2pt]
\hspace*{4.0em}\lstinline[basicstyle=\ttfamily\footnotesize,columns=fullflexible,keepspaces=true,showstringspaces=false,breaklines=true,breakatwhitespace=false]|- Monotonicity vs non-monotonicity tradeoffs|\tabularnewline[-2pt]
\hspace*{4.0em}\lstinline[basicstyle=\ttfamily\footnotesize,columns=fullflexible,keepspaces=true,showstringspaces=false,breaklines=true,breakatwhitespace=false]|- Smoothness (differentiability everywhere vs piecewise)|\tabularnewline[-2pt]
\hspace*{4.0em}\lstinline[basicstyle=\ttfamily\footnotesize,columns=fullflexible,keepspaces=true,showstringspaces=false,breaklines=true,breakatwhitespace=false]|- Negative-domain behavior (zero, linear, bounded)|\tabularnewline[-2pt]
\hspace*{4.0em}\lstinline[basicstyle=\ttfamily\footnotesize,columns=fullflexible,keepspaces=true,showstringspaces=false,breaklines=true,breakatwhitespace=false]|- Computational cost (simple ops vs learned parameters)|\tabularnewline[-2pt]
\hspace*{4.0em}\lstinline[basicstyle=\ttfamily\footnotesize,columns=fullflexible,keepspaces=true,showstringspaces=false,breaklines=true,breakatwhitespace=false]|- Interaction with BatchNorm and residual connections|\tabularnewline[-2pt]
\hspace*{2.0em}\lstinline[basicstyle=\ttfamily\footnotesize,columns=fullflexible,keepspaces=true,showstringspaces=false,breaklines=true,breakatwhitespace=false]|"""|\tabularnewline[-2pt]
\rule{0pt}{4pt}\tabularnewline[-4pt]
\hspace*{2.0em}\lstinline[basicstyle=\ttfamily\footnotesize,columns=fullflexible,keepspaces=true,showstringspaces=false,breaklines=true,breakatwhitespace=false]|def __init__(self):|\tabularnewline[-2pt]
\hspace*{4.0em}\lstinline[basicstyle=\ttfamily\footnotesize,columns=fullflexible,keepspaces=true,showstringspaces=false,breaklines=true,breakatwhitespace=false]|super().__init__()|\tabularnewline[-2pt]
\rule{0pt}{4pt}\tabularnewline[-4pt]
\hspace*{2.0em}\lstinline[basicstyle=\ttfamily\footnotesize,columns=fullflexible,keepspaces=true,showstringspaces=false,breaklines=true,breakatwhitespace=false]|def forward(self, x):|\tabularnewline[-2pt]
\hspace*{4.0em}\lstinline[basicstyle=\ttfamily\footnotesize,columns=fullflexible,keepspaces=true,showstringspaces=false,breaklines=true,breakatwhitespace=false]|return F.relu(x)  # default: ReLU|\tabularnewline[-2pt]
\end{longtable}
\endgroup

\begingroup
\footnotesize
\setlength{\LTleft}{0pt}
\setlength{\LTright}{0pt}
\setlength{\LTcapwidth}{\linewidth}
\setlength{\LTpre}{6pt}
\setlength{\LTpost}{10pt}
\renewcommand{\arraystretch}{1.08}
\begin{longtable}{@{}>{\raggedright\arraybackslash}p{\linewidth}@{}}
\caption{Idea proposal for the activation-function example. The system instruction, stage-specific user instruction, and recorded approach are shown. Repeated task and parent inputs are referenced by table number.}\label{tab:pipeline-proposal}\\
\toprule
\textbf{Stage 1: idea proposal}\tabularnewline
\endfirsthead
\multicolumn{1}{@{}l@{}}{\footnotesize\tablename~\thetable{} (continued)}\\
\toprule
\textbf{Stage 1: idea proposal}\tabularnewline
\midrule
\endhead
\midrule
\multicolumn{1}{@{}r@{}}{\footnotesize\emph{Continued on the next page.}}\\
\endfoot
\bottomrule
\endlastfoot
\midrule
\textbf{System prompt}\tabularnewline*[3pt]
You are an expert ML researcher proposing ONE concrete improvement to an existing solution. You propose the IDEA only \textemdash{} you do NOT write code in this step. Be specific and technical: name the modeling/training/feature change and why it should move the target metric. Propose a single coherent idea, not a menu of options.\tabularnewline[3pt]
\midrule
\textbf{User prompt}\tabularnewline*[3pt]
\textbf{Task: mls\_\_dl-activation-function}\tabularnewline*[2pt]
\emph{Shared input: insert the complete task description from Table~\ref{tab:pipeline-task-parent}.}\tabularnewline[3pt]
Metric: \textbf{combined\_score} (higher is better)\tabularnewline[3pt]
\textbf{Current parent solution (current parent verifier score: 0.5997 \textemdash{} beat it)}\tabularnewline*[2pt]
\emph{Shared input: insert the parent editable band from Table~\ref{tab:pipeline-task-parent}.}\tabularnewline[3pt]
\textbf{Inspiration: an alternative approach (verifier score: 0.5997)}\tabularnewline*[2pt]
A neutral summary of a DIFFERENT approach that also scored well. Use it as a source of ideas to combine/contrast/diverge from \textemdash{} do NOT copy it verbatim. (baseline)\tabularnewline[3pt]
\textbf{Your task}\tabularnewline*[2pt]
Propose ONE concrete, self-contained improvement idea over the parent solution above that should improve the metric (higher is better). It must be implementable as an edit to the parent within the runtime budget, using only standard ML libraries and the data already present \textemdash{} do NOT invent data, files, columns, or external resources.\tabularnewline[3pt]
Emit EXACTLY two sections and nothing else:\tabularnewline[3pt]
\textbf{Approach}\tabularnewline*[2pt]
The concrete change(s) you propose and the mechanism by which they should improve the metric. Be specific (architecture/algorithm/hyperparameters/features/loss/validation), not vague.>\tabularnewline[3pt]
\midrule
\textbf{Proposed approach (actual model output, source Section 1.3)}\tabularnewline*[3pt]
We propose \textbf{Mish-ACON-C}, a parameterized activation function that merges the Activate Or Not (ACON) framework with the Mish activation to learn layer-specific non-linearities. The function computes \lstinline[basicstyle=\ttfamily\footnotesize,columns=fullflexible,breaklines=true,breakatwhitespace=false]|diff = (p1 - p2) * x| and outputs \lstinline[basicstyle=\ttfamily\footnotesize,columns=fullflexible,breaklines=true,breakatwhitespace=false]|diff * torch.tanh(F.softplus(beta * diff)) + p2 * x|, governed by three learnable scalar parameters per layer: positive slope (\lstinline[basicstyle=\ttfamily\footnotesize,columns=fullflexible,breaklines=true,breakatwhitespace=false]|p1|), negative slope (\lstinline[basicstyle=\ttfamily\footnotesize,columns=fullflexible,breaklines=true,breakatwhitespace=false]|p2|), and transition sharpness (\lstinline[basicstyle=\ttfamily\footnotesize,columns=fullflexible,breaklines=true,breakatwhitespace=false]|beta|). To guarantee stability under the fixed SGD optimizer and prevent weight decay from collapsing the slopes, these scalars are formulated as zero-initialized offsets bounded by a \lstinline[basicstyle=\ttfamily\footnotesize,columns=fullflexible,breaklines=true,breakatwhitespace=false]|tanh| squash (e.g., \lstinline[basicstyle=\ttfamily\footnotesize,columns=fullflexible,breaklines=true,breakatwhitespace=false]|p1 = 1.0 + 0.5 * torch.tanh(p1_offset)|, \lstinline[basicstyle=\ttfamily\footnotesize,columns=fullflexible,breaklines=true,breakatwhitespace=false]|p2 = 0.25 * torch.tanh(p2_offset)|, \lstinline[basicstyle=\ttfamily\footnotesize,columns=fullflexible,breaklines=true,breakatwhitespace=false]|beta = 1.0 + 0.8 * torch.tanh(beta_offset)|). By defaulting exactly to the state-of-the-art Mish baseline at initialization, this mechanism strictly expands the function space safely, allowing the network to dynamically tune its gating and asymptotes to improve test accuracy without risking gradient explosion.\tabularnewline[3pt]
\end{longtable}
\endgroup

\newpage
\begingroup
\footnotesize
\setlength{\LTleft}{0pt}
\setlength{\LTright}{0pt}
\setlength{\LTcapwidth}{\linewidth}
\setlength{\LTpre}{6pt}
\setlength{\LTpost}{10pt}
\renewcommand{\arraystretch}{1.08}
\begin{longtable}{@{}>{\raggedright\arraybackslash}p{\linewidth}@{}}
\caption{Implementation of the committed Mish-ACON-C idea. The replacement code is produced before critic model screening and is subsequently reused for verification. Repeated inputs and the full fixed code context are referenced by table number.}\label{tab:pipeline-implementation}\\
\toprule
\textbf{Stage 2: implementation}\tabularnewline
\endfirsthead
\multicolumn{1}{@{}l@{}}{\footnotesize\tablename~\thetable{} (continued)}\\
\toprule
\textbf{Stage 2: implementation}\tabularnewline
\midrule
\endhead
\midrule
\multicolumn{1}{@{}r@{}}{\footnotesize\emph{Continued on the next page.}}\\
\endfoot
\bottomrule
\endlastfoot
\midrule
\textbf{System prompt}\tabularnewline*[3pt]
You are an expert ML researcher. Implement the committed idea by rewriting ONLY the editable band of the file. Emit the replacement band in a single ``\lstinline[basicstyle=\ttfamily\footnotesize,columns=fullflexible,breaklines=true,breakatwhitespace=false]|python|`` block and nothing else after it.\tabularnewline[3pt]
\midrule
\textbf{User prompt}\tabularnewline*[3pt]
\textbf{Committed idea (implement EXACTLY this)}\tabularnewline*[2pt]
You have already decided on the following idea. Implement EXACTLY this idea. Do NOT change the approach, do NOT substitute a different method, do NOT add unrelated changes \textemdash{} just turn this idea into correct, runnable code.\tabularnewline[3pt]
\emph{Shared input: insert the complete proposed approach from Table~\ref{tab:pipeline-proposal}.}\tabularnewline[3pt]
\textbf{Task: dl-activation-function}\tabularnewline*[2pt]
\emph{Shared input: insert the complete task description from Table~\ref{tab:pipeline-task-parent}.}\tabularnewline[3pt]
\textbf{Editable contract}\tabularnewline*[2pt]
You may ONLY replace the EDITABLE band (lines 32-49) of \lstinline[basicstyle=\ttfamily\footnotesize,columns=fullflexible,breaklines=true,breakatwhitespace=false]|pytorch-vision/custom_activation.py|. Everything OUTSIDE that band (the FIXED header/footer shown below, and any read-only sibling files) is IMMUTABLE \textemdash{} the official verifier byte-compares it against the reference and scores 0 if it changed. Your replacement must keep the SAME public interface (function/class/variable names, signatures) that the FIXED code imports or calls.\tabularnewline[3pt]
\textbf{Current editable band (lines 32-49)}\tabularnewline*[2pt]
\emph{Shared input: insert the parent editable band from Table~\ref{tab:pipeline-task-parent}.}\tabularnewline[3pt]
Return the FULL replacement for this band (a drop-in for lines 32-49) in one ``\lstinline[basicstyle=\ttfamily\footnotesize,columns=fullflexible,breaklines=true,breakatwhitespace=false]|python|`` block.\tabularnewline[3pt]
\midrule
\textbf{Implemented child code}\tabularnewline*[3pt]
\lstinline[basicstyle=\ttfamily\footnotesize,columns=fullflexible,keepspaces=true,showstringspaces=false,breaklines=true,breakatwhitespace=false]|class CustomActivation(nn.Module):|\tabularnewline[-2pt]
\hspace*{2.0em}\lstinline[basicstyle=\ttfamily\footnotesize,columns=fullflexible,keepspaces=true,showstringspaces=false,breaklines=true,breakatwhitespace=false]|"""Custom activation function. Drop-in replacement for ReLU.|\tabularnewline[-2pt]
\rule{0pt}{4pt}\tabularnewline[-4pt]
\hspace*{2.0em}\lstinline[basicstyle=\ttfamily\footnotesize,columns=fullflexible,keepspaces=true,showstringspaces=false,breaklines=true,breakatwhitespace=false]|Input/Output: tensor of any shape, element-wise operation.|\tabularnewline[-2pt]
\rule{0pt}{4pt}\tabularnewline[-4pt]
\hspace*{2.0em}\lstinline[basicstyle=\ttfamily\footnotesize,columns=fullflexible,keepspaces=true,showstringspaces=false,breaklines=true,breakatwhitespace=false]|Design considerations:|\tabularnewline[-2pt]
\hspace*{4.0em}\lstinline[basicstyle=\ttfamily\footnotesize,columns=fullflexible,keepspaces=true,showstringspaces=false,breaklines=true,breakatwhitespace=false]|- Monotonicity vs non-monotonicity tradeoffs|\tabularnewline[-2pt]
\hspace*{4.0em}\lstinline[basicstyle=\ttfamily\footnotesize,columns=fullflexible,keepspaces=true,showstringspaces=false,breaklines=true,breakatwhitespace=false]|- Smoothness (differentiability everywhere vs piecewise)|\tabularnewline[-2pt]
\hspace*{4.0em}\lstinline[basicstyle=\ttfamily\footnotesize,columns=fullflexible,keepspaces=true,showstringspaces=false,breaklines=true,breakatwhitespace=false]|- Negative-domain behavior (zero, linear, bounded)|\tabularnewline[-2pt]
\hspace*{4.0em}\lstinline[basicstyle=\ttfamily\footnotesize,columns=fullflexible,keepspaces=true,showstringspaces=false,breaklines=true,breakatwhitespace=false]|- Computational cost (simple ops vs learned parameters)|\tabularnewline[-2pt]
\hspace*{4.0em}\lstinline[basicstyle=\ttfamily\footnotesize,columns=fullflexible,keepspaces=true,showstringspaces=false,breaklines=true,breakatwhitespace=false]|- Interaction with BatchNorm and residual connections|\tabularnewline[-2pt]
\hspace*{2.0em}\lstinline[basicstyle=\ttfamily\footnotesize,columns=fullflexible,keepspaces=true,showstringspaces=false,breaklines=true,breakatwhitespace=false]|"""|\tabularnewline[-2pt]
\rule{0pt}{4pt}\tabularnewline[-4pt]
\hspace*{2.0em}\lstinline[basicstyle=\ttfamily\footnotesize,columns=fullflexible,keepspaces=true,showstringspaces=false,breaklines=true,breakatwhitespace=false]|def __init__(self):|\tabularnewline[-2pt]
\hspace*{4.0em}\lstinline[basicstyle=\ttfamily\footnotesize,columns=fullflexible,keepspaces=true,showstringspaces=false,breaklines=true,breakatwhitespace=false]|super().__init__()|\tabularnewline[-2pt]
\hspace*{4.0em}\lstinline[basicstyle=\ttfamily\footnotesize,columns=fullflexible,keepspaces=true,showstringspaces=false,breaklines=true,breakatwhitespace=false]|self.p1_offset = nn.Parameter(torch.zeros(1))|\tabularnewline[-2pt]
\hspace*{4.0em}\lstinline[basicstyle=\ttfamily\footnotesize,columns=fullflexible,keepspaces=true,showstringspaces=false,breaklines=true,breakatwhitespace=false]|self.p2_offset = nn.Parameter(torch.zeros(1))|\tabularnewline[-2pt]
\hspace*{4.0em}\lstinline[basicstyle=\ttfamily\footnotesize,columns=fullflexible,keepspaces=true,showstringspaces=false,breaklines=true,breakatwhitespace=false]|self.beta_offset = nn.Parameter(torch.zeros(1))|\tabularnewline[-2pt]
\rule{0pt}{4pt}\tabularnewline[-4pt]
\hspace*{2.0em}\lstinline[basicstyle=\ttfamily\footnotesize,columns=fullflexible,keepspaces=true,showstringspaces=false,breaklines=true,breakatwhitespace=false]|def forward(self, x):|\tabularnewline[-2pt]
\hspace*{4.0em}\lstinline[basicstyle=\ttfamily\footnotesize,columns=fullflexible,keepspaces=true,showstringspaces=false,breaklines=true,breakatwhitespace=false]|p1 = 1.0 + 0.5 * torch.tanh(self.p1_offset)|\tabularnewline[-2pt]
\hspace*{4.0em}\lstinline[basicstyle=\ttfamily\footnotesize,columns=fullflexible,keepspaces=true,showstringspaces=false,breaklines=true,breakatwhitespace=false]|p2 = 0.25 * torch.tanh(self.p2_offset)|\tabularnewline[-2pt]
\hspace*{4.0em}\lstinline[basicstyle=\ttfamily\footnotesize,columns=fullflexible,keepspaces=true,showstringspaces=false,breaklines=true,breakatwhitespace=false]|beta = 1.0 + 0.8 * torch.tanh(self.beta_offset)|\tabularnewline[-2pt]
\rule{0pt}{4pt}\tabularnewline[-4pt]
\hspace*{4.0em}\lstinline[basicstyle=\ttfamily\footnotesize,columns=fullflexible,keepspaces=true,showstringspaces=false,breaklines=true,breakatwhitespace=false]|diff = (p1 - p2) * x|\tabularnewline[-2pt]
\hspace*{4.0em}\lstinline[basicstyle=\ttfamily\footnotesize,columns=fullflexible,keepspaces=true,showstringspaces=false,breaklines=true,breakatwhitespace=false]|return diff * torch.tanh(F.softplus(beta * diff)) + p2 * x|\tabularnewline[-2pt]
\end{longtable}
\endgroup

\begingroup
\footnotesize
\setlength{\LTleft}{0pt}
\setlength{\LTright}{0pt}
\setlength{\LTcapwidth}{\linewidth}
\setlength{\LTpre}{6pt}
\setlength{\LTpost}{10pt}
\renewcommand{\arraystretch}{1.08}
\begin{longtable}{@{}>{\raggedright\arraybackslash}p{\linewidth}@{}}
\caption{Code-grounded summarization of the implemented child. The summarizer receives the task, proposed approach, and implemented code; its recorded output forms the child description supplied to the critic model.}\label{tab:pipeline-summarization}\\
\toprule
\textbf{Stage 3: solution summarization}\tabularnewline
\endfirsthead
\multicolumn{1}{@{}l@{}}{\footnotesize\tablename~\thetable{} (continued)}\\
\toprule
\textbf{Stage 3: solution summarization}\tabularnewline
\midrule
\endhead
\midrule
\multicolumn{1}{@{}r@{}}{\footnotesize\emph{Continued on the next page.}}\\
\endfoot
\bottomrule
\endlastfoot
\midrule
\textbf{System prompt}\tabularnewline*[3pt]
You are an expert ML solver. You are given a task, a candidate solution's code, and its short approach. Write a NEUTRAL, purely descriptive summary of what the solution does. Do NOT evaluate, rank, judge, compare, or predict scores.\tabularnewline[3pt]
\midrule
\textbf{User prompt}\tabularnewline*[3pt]
\textbf{Task: mls\_\_dl-activation-function}\tabularnewline*[2pt]
\emph{Shared input: insert the complete task description from Table~\ref{tab:pipeline-task-parent}.}\tabularnewline[3pt]
Metric: \textbf{combined\_score} (higher is better)\tabularnewline[3pt]
\textbf{Approach}\tabularnewline*[2pt]
\emph{Shared input: insert the complete proposed approach from Table~\ref{tab:pipeline-proposal}.}\tabularnewline[3pt]
\textbf{Code}\tabularnewline*[2pt]
\emph{Shared input: insert the complete implemented child code from Table~\ref{tab:pipeline-implementation}.}\tabularnewline[3pt]
\textbf{Your task}\tabularnewline*[2pt]
Write a 120-200 word NEUTRAL descriptive summary of what the code does: task framing, model/algorithm, training/eval setup, and task-specific handling. PURELY DESCRIPTIVE \textemdash{} do NOT evaluate, rank, compare, or guess the score. Output only the summary.\tabularnewline[3pt]
\midrule
\textbf{Child summary}\tabularnewline*[3pt]
This solution implements a custom activation function for deep convolutional neural networks (ResNet, VGG, MobileNetV2) evaluated on CIFAR-10, CIFAR-100, and FashionMNIST. The approach, termed Mish-ACON-C, integrates the Activate Or Not (ACON) framework with the Mish activation function to create a parameterized, layer-specific non-linearity. The \lstinline[basicstyle=\ttfamily\footnotesize,columns=fullflexible,breaklines=true,breakatwhitespace=false]|CustomActivation| module defines three learnable scalar parameters per layer: \lstinline[basicstyle=\ttfamily\footnotesize,columns=fullflexible,breaklines=true,breakatwhitespace=false]|p1_offset|, \lstinline[basicstyle=\ttfamily\footnotesize,columns=fullflexible,breaklines=true,breakatwhitespace=false]|p2_offset|, and \lstinline[basicstyle=\ttfamily\footnotesize,columns=fullflexible,breaklines=true,breakatwhitespace=false]|beta_offset|, all initialized to zero. To maintain stability during training with the fixed SGD optimizer and restrict weight decay collapse, these offsets are passed through a \lstinline[basicstyle=\ttfamily\footnotesize,columns=fullflexible,breaklines=true,breakatwhitespace=false]|tanh| function to create bounded variables: a positive slope (\lstinline[basicstyle=\ttfamily\footnotesize,columns=fullflexible,breaklines=true,breakatwhitespace=false]|p1 = 1.0 + 0.5 * tanh(p1_offset)|), a negative slope (\lstinline[basicstyle=\ttfamily\footnotesize,columns=fullflexible,breaklines=true,breakatwhitespace=false]|p2 = 0.25 * tanh(p2_offset)|), and a transition sharpness (\lstinline[basicstyle=\ttfamily\footnotesize,columns=fullflexible,breaklines=true,breakatwhitespace=false]|beta = 1.0 + 0.8 * tanh(beta_offset)|).\tabularnewline[3pt]
During the forward pass, the module computes a scaled input \lstinline[basicstyle=\ttfamily\footnotesize,columns=fullflexible,breaklines=true,breakatwhitespace=false]|diff = (p1 - p2) * x| and applies the transformation \lstinline[basicstyle=\ttfamily\footnotesize,columns=fullflexible,breaklines=true,breakatwhitespace=false]|diff * tanh(softplus(beta * diff)) + p2 * x|. Because the offsets are zero-initialized, the function initially behaves exactly like the baseline Mish activation (\lstinline[basicstyle=\ttfamily\footnotesize,columns=fullflexible,breaklines=true,breakatwhitespace=false]|x * tanh(softplus(x))|), allowing the network to adjust its gating and asymptotes throughout the training process.\tabularnewline[3pt]
\end{longtable}
\endgroup

\subsection{Task understanding.}\label{app:task-understanding}
Table~\ref{tab:pipeline-task-understanding} presents the precomputed task context supplied to the critic model. It describes the evaluation objective, submission interface, datasets, input shapes, and preprocessing. This context is generated once per task and reused across candidates.

\newpage
\begingroup
\footnotesize
\setlength{\LTleft}{0pt}
\setlength{\LTright}{0pt}
\setlength{\LTcapwidth}{\linewidth}
\setlength{\LTpre}{6pt}
\setlength{\LTpost}{10pt}
\renewcommand{\arraystretch}{1.08}
\begin{longtable}{@{}>{\raggedright\arraybackslash}p{\linewidth}@{}}
\caption{Precomputed task understanding supplied to the critic model for activation-function design. This context is generated once per task and reused across candidates; it is shown here in its recorded wording.}\label{tab:pipeline-task-understanding}\\
\toprule
\textbf{Task-specific context for the critic model}\tabularnewline
\endfirsthead
\multicolumn{1}{@{}l@{}}{\footnotesize\tablename~\thetable{} (continued)}\\
\toprule
\textbf{Task-specific context for the critic model}\tabularnewline
\midrule
\endhead
\midrule
\multicolumn{1}{@{}r@{}}{\footnotesize\emph{Continued on the next page.}}\\
\endfoot
\bottomrule
\endlastfoot
\midrule
\textbf{Task understanding}\tabularnewline*[3pt]
\textbf{Modality:} Computer Vision (Image Classification)\tabularnewline[3pt]
\textbf{Metric:} Test accuracy (higher-is-better)\tabularnewline[3pt]
\textbf{Overview}\tabularnewline*[2pt]
The task requires implementing a custom activation function as a PyTorch \lstinline[basicstyle=\ttfamily\footnotesize,columns=fullflexible,breaklines=true,breakatwhitespace=false]|nn.Module| to serve as a drop-in replacement for standard activations in ResNet, VGG, and MobileNetV2 architectures. The submission format is Python code modifying the \lstinline[basicstyle=\ttfamily\footnotesize,columns=fullflexible,breaklines=true,breakatwhitespace=false]|forward| method and optionally the \lstinline[basicstyle=\ttfamily\footnotesize,columns=fullflexible,breaklines=true,breakatwhitespace=false]|__init__| method of the \lstinline[basicstyle=\ttfamily\footnotesize,columns=fullflexible,breaklines=true,breakatwhitespace=false]|CustomActivation| class within an 18-line editable region in \lstinline[basicstyle=\ttfamily\footnotesize,columns=fullflexible,breaklines=true,breakatwhitespace=false]|pytorch-vision/custom_activation.py|. The implemented function must perform an element-wise or channel-wise operation that preserves the exact shape of the input tensor.\tabularnewline[3pt]
\textbf{Data}\tabularnewline*[2pt]
* \textbf{Source Datasets:} CIFAR-10, CIFAR-100, and FashionMNIST.\tabularnewline[2pt]
* \textbf{Target Cardinality:} 10 classes for CIFAR-10 and FashionMNIST; 100 classes for CIFAR-100.\tabularnewline[2pt]
* \textbf{Input Image Dimensions:} All dataset images are resized to 32x32 pixels.\tabularnewline[2pt]
* \textbf{Channels:} CIFAR-10 and CIFAR-100 are natively 3-channel RGB. FashionMNIST is natively grayscale but is explicitly repeated across 3 channels (\lstinline[basicstyle=\ttfamily\footnotesize,columns=fullflexible,breaklines=true,breakatwhitespace=false]|x.repeat(3, 1, 1)|) in the data pipeline.\tabularnewline[2pt]
* \textbf{Activation Tensor Shapes:} The custom activation receives intermediate feature map tensors of varying shapes depending on the specific architecture (ResNet-20, ResNet-56, VGG-16, MobileNetV2) and layer depth.\tabularnewline[2pt]
* \textbf{Batch Size:} 128 for both training and testing dataloaders.\tabularnewline[2pt]
* \textbf{Normalization Statistics:}\tabularnewline[2pt]
* CIFAR-10: mean \lstinline[basicstyle=\ttfamily\footnotesize,columns=fullflexible,breaklines=true,breakatwhitespace=false]|(0.4914, 0.4822, 0.4465)|, std \lstinline[basicstyle=\ttfamily\footnotesize,columns=fullflexible,breaklines=true,breakatwhitespace=false]|(0.2023, 0.1994, 0.2010)|.\tabularnewline[2pt]
* CIFAR-100: mean \lstinline[basicstyle=\ttfamily\footnotesize,columns=fullflexible,breaklines=true,breakatwhitespace=false]|(0.5071, 0.4867, 0.4408)|, std \lstinline[basicstyle=\ttfamily\footnotesize,columns=fullflexible,breaklines=true,breakatwhitespace=false]|(0.2675, 0.2565, 0.2761)|.\tabularnewline[2pt]
* FashionMNIST: mean \lstinline[basicstyle=\ttfamily\footnotesize,columns=fullflexible,breaklines=true,breakatwhitespace=false]|(0.2860, 0.2860, 0.2860)|, std \lstinline[basicstyle=\ttfamily\footnotesize,columns=fullflexible,breaklines=true,breakatwhitespace=false]|(0.3530, 0.3530, 0.3530)|.\tabularnewline[2pt]
* \textbf{Data Augmentation:} Training sets apply \lstinline[basicstyle=\ttfamily\footnotesize,columns=fullflexible,breaklines=true,breakatwhitespace=false]|RandomCrop(32, padding=4)| and \lstinline[basicstyle=\ttfamily\footnotesize,columns=fullflexible,breaklines=true,breakatwhitespace=false]|RandomHorizontalFlip()|. Both train and test sets apply \lstinline[basicstyle=\ttfamily\footnotesize,columns=fullflexible,breaklines=true,breakatwhitespace=false]|ToTensor()| and normalization.\tabularnewline[2pt]
\end{longtable}
\endgroup

\section{Complete Prompt Instructions}
\label{app:prompts}

Table~\ref{tab:critic_prompt} presents the complete input and output format used for critic training and inference.

\begin{table*}[t]
\centering
\small
\setlength{\tabcolsep}{6pt}
\renewcommand{\arraystretch}{1.15}
\caption{Prompt template for the idea-level critic. The provided
task-specific criteria must all be considered, while the critic may
freely determine their relative importance and aggregation.}
\label{tab:critic_prompt}

\begin{tabularx}{\textwidth}{
    @{}
    >{\raggedright\arraybackslash\bfseries}p{0.15\textwidth}
    >{\raggedright\arraybackslash}X
    @{}
}
\toprule
Component & Prompt content \\
\midrule

Role
&
You are an expert ML solution critic. Given a parent solution and a
candidate child idea, predict whether implementing the child idea will
improve the task metric over the parent.
\\
\midrule

Decision stages
&
\textbf{Stage 1: Improvement.}
Determine whether the child will \texttt{improve}, \texttt{lose}, or
\texttt{tie} relative to the parent.

\textbf{Stage 2: Win tier.}
If the Stage-1 prediction is \texttt{improve}, determine whether the
expected gain is \texttt{big}, \texttt{medium}, or \texttt{small}.
Otherwise, output \texttt{NA}.
\\
\midrule

Task input
&
\texttt{\{task\_description\}}\\

\textbf{\# Parent (baseline)} & \texttt{\{parent\_idea\_summary\}}\\

\textbf{\# Child (candidate)} & \texttt{\{child\_idea\_summary\}}
\\
\midrule

Task context
&
\texttt{\{neutral\_understanding\_md\}}\\
\textbf{Task-specific criteria} & \texttt{\{task\_specific\_criteria\_md\}}
\\
\midrule

Soft-rubric
reasoning
&
Assess the child idea against the parent solution using the provided
task-specific criteria.

\begin{enumerate}
    \setlength{\itemsep}{1pt}
    \setlength{\topsep}{3pt}
    \item Evaluate every provided criterion. You may introduce an
    additional criterion only when it captures an important factor
    missing from the provided rubric.
    \item For each criterion, state whether the child is
    \texttt{better}, \texttt{worse}, or \texttt{about the same} as the
    parent.
    \item Determine the importance of each criterion for the current
    task, using either qualitative weights
    (\texttt{high}/\texttt{medium}/\texttt{low}) or approximate
    percentages, and briefly justify the weighting.
    \item Explain how the weighted criterion-level comparisons are
    combined into the final prediction.
\end{enumerate}

The criteria constrain what must be evaluated, but the analysis,
weighting, and aggregation strategy remain flexible.
\\
\midrule

Output format
&
\texttt{<criterion-level comparisons and explicit weighted aggregation>} \\

\textbf{\# Verdict} &
\texttt{[IMPROVEMENT] improve | lose | tie --- <one-sentence reason>}
\texttt{[WINTIER] big | medium | small | NA --- <one-sentence reason>}
\\
\midrule

Output constraints
&
Use the exact tags \texttt{[IMPROVEMENT]} and \texttt{[WINTIER]}.
Provide a calibrated and decisive prediction. The
\texttt{[WINTIER]} prediction must be \texttt{NA} whenever the
\texttt{[IMPROVEMENT]} prediction is \texttt{lose} or \texttt{tie}.
\\
\bottomrule
\end{tabularx}
\end{table*}

\section{Soft-Rubric Reasoning Examples}
\label{app:reasoning_examples}

Table~\ref{tab:reasoning_example} presents example critiques with freestyle and soft-rubric reasoning.
\begin{table*}[t]
\centering
\footnotesize
\setlength{\tabcolsep}{5pt}
\renewcommand{\arraystretch}{1.12}
\caption{Qualitative comparison between freestyle and soft
rubric-based critic reasoning. The example is abridged for readability.
Freestyle reasoning focuses primarily on the risk of overfitting,
whereas soft rubric reasoning explicitly compares the parent and child
under task-specific criteria before aggregating the judgments.}
\label{tab:reasoning_example}

\begin{tabularx}{\textwidth}{
    @{}
    >{\raggedright\arraybackslash\bfseries}p{0.09\textwidth}
    >{\hsize=0.65\hsize\linewidth=\hsize
      \raggedright\arraybackslash}X
    >{\hsize=1.35\hsize\linewidth=\hsize
      \raggedright\arraybackslash}X
    @{}
}
\toprule
&
\multicolumn{1}{c}{\textbf{Freestyle Reasoning}}
&
\multicolumn{1}{c}{\textbf{Soft Rubric-Based Reasoning}}
\\
\midrule

Analysis
&
The child proposes a custom U-Net trained from scratch on a subset of the available data, with images downsampled to $128\times128$. Although the pipeline is computationally feasible, training a CNN from scratch on a small medical-imaging dataset without pretrained weights or data augmentation is likely to cause severe overfitting.

\medskip
The parent instead relies on statistical priors, such as mean masks and spatial probabilities. These priors are robust on small datasets and can exploit the relatively consistent anatomical locations of the target organs. The freestyle critic therefore predicts that the child will generalize worse than the parent.
&
\textbf{1. Criteria considered.}

\emph{MRI handling and feature extraction:} whether the method captures spatial structure, intensity patterns, and patient-specific anatomical variation.

\emph{Dice and Hausdorff performance:} whether the method can produce accurate masks and boundaries under the task metric.
...
\medskip

\textbf{2. Criterion importance.}

Feature extraction receives high importance ($40\%$) because medical segmentation depends on learning spatial and intensity features.
Metric alignment also receives high importance ($30\%$), particularly because Hausdorff distance is sensitive to boundary errors and outlier pixels.
...
\medskip

\textbf{3. Parent--child comparison.}

For feature extraction, the child is substantially better: the parent's statistical templates cannot capture patient-specific variation, whereas U-Net can learn nonlinear hierarchical features directly from the images.
For metric alignment, the child is also better: U-Net's skip connections can produce more accurate and continuous segmentation boundaries than fixed statistical priors.
....
\medskip

\textbf{4. Aggregation.}

After weighting the criterion-level comparisons, the expected benefits of learned spatial features and improved boundaries outweigh the child's increased risk of overfitting.
\\
\midrule

Verdict
&
\texttt{[IMPROVEMENT]} \textbf{lose}

The child is likely to overfit because it trains a U-Net from scratch
on only a subset of a small dataset.

\medskip
\texttt{[WINTIER]} \textbf{NA}

The predicted outcome is not an improvement.
&
\texttt{[IMPROVEMENT]} \textbf{improve}

The child replaces a rigid statistical prior with a model capable of
learning task-specific spatial and intensity features.

\medskip
\texttt{[WINTIER]} \textbf{big}

Moving from a non-learned statistical baseline to U-Net is expected to
produce a substantial improvement in both mask accuracy and boundary
quality.
\\
\bottomrule
\end{tabularx}
\end{table*}

\section{Experiment Setup Details}
\label{app:training_details}

\subsection{Supervised Fine-Tuning Configuration}

We train both critic model sizes for three epochs using Low-Rank Adaptation (LoRA) with rank 32. We use a learning rate of $1\times10^{-5}$ for all SFT experiments and run for 3 epochs. 

\subsection{GRPO Configuration}

For both 9B and 35B models, starting from the SFT checkpoints, we run GRPO for 300 rollout steps, with each step processing 256 idea inputs. For each idea, the critic generates a group of eight responses. We discard groups in which all responses receive the same reward, as these groups contain no relative learning signal for GRPO. We use the temperature of 0.7, learning rate of $1 \times 10^{-6}$, weight decay of 0.1, and KL coefficient of 0.01. 

\subsection{Policy GRPO Training Details}

We initialize the idea-proposing policy from both Qwen-3.5-9B and Qwen-3.5-35B and train it on 95 MLE-Dojo and 46 MLS-Bench environments. For critic-guided reward assignment, we use the frozen Critic-9B and Critic-35B checkpoint obtained after SFT and GRPO for training idea proposer Qwen-3.5-9B and Qwen-3.5-35B respectfully. Each training step collects 16 parent-conditioned groups, with 16 candidate ideas per group. The initial parent for each task is its seed program. Whenever all 16 candidates in a group receive an \textsc{Improve} verdict (i.e., an assigned reward of $+1$), we randomly select one child to become the task's new parent; otherwise, the parent remains unchanged. Each method receives a total budget of 10,000 ground-truth verifications across all training environments.

The policy generates only candidate ideas, while Gemini-3.1-Pro is responsible for implementing ideas and summarizing implementations. If a solution is selected to be verified, for non-timeout execution errors, Gemini-3.1-Pro is allowed
up to three repair attempts. Both the initial implementation and subsequent repairs must strictly follow the proposed idea: Gemini-3.1-Pro is instructed to resolve implementation errors without introducing new algorithmic choices or modifying the substantive proposal.

For both 9B and 35B models, the critic-in-loop checkpoints are trained for around 110 steps while the verify-all checkpoints are trained for around 40 steps, targeting the 10,000 verification budget. We use learning rate of $1\times 10^-6$ and weight decay of 0.1. The temperature used is 0.7. 

\section{Additional Critic Evaluation Results}
\label{app:additional_results}

\subsection{Full Freestyle and Soft-Rubric Ablation Results}

Table~\ref{tab:critic_freestyle} shows the results for critique using freestyle reasoning vs. soft rubric reasoning with all metrics. 

\begin{table*}[h]
\centering
\small
\setlength{\tabcolsep}{5pt}
\caption{Idea-critic performance on the 1.5K-example evaluation set. SFT substantially improves both model sizes, while criterion-guided GRPO provides consistent additional gains. Best results are shown in bold.}
\label{tab:critic_freestyle}
\begin{tabular}{llcccc}
\toprule
\textbf{Model} & \textbf{Training }
& \textbf{Acc}
& \textbf{Dir-3}
& \textbf{F1}
& \textbf{5-Ladder} \\
\midrule
Qwen-3.5-9B         & Base                       & 49.9 & 41.3 & 58.3 & 22.4 \\
Qwen-3.5-35B        & Base                       & 49.9 & 36.4 & 51.8 & 20.1 \\
\midrule
Qwen-3.5-9B         & SFT, freestyle             & 61.0 & 54.6 & 71.5 & 31.9 \\
Qwen-3.5-9B         & SFT, soft-rubric           & 62.4 & 55.6 & 70.9 & 31.5 \\
\midrule
Qwen-3.5-9B         & SFT+GRPO, freestyle        & 49.1 & 40.3 & 48.6 & 26.8 \\
Qwen-3.5-9B         & SFT+GRPO, soft-rubric      & 66.7 & 60.7 & \textbf{75.6} & \textbf{41.5} \\
\bottomrule
\end{tabular}
\end{table*}

\subsection{Results across Sampled, Revised, and Guided Examples}

Table~\ref{tab:difficulty_breakdown} shows the detailed breakdown of results in Table~\ref{tab:critic_main}. We show the results across the Sampled, Revised, and Guided subsets separately.
\begin{table*}[t]
\centering
\scriptsize
\setlength{\tabcolsep}{1.5pt}
\renewcommand{\arraystretch}{1.12}
\caption{Critic performance (\%) across the Sampled, Revised, and
Guided subsets. These subsets correspond to increasing levels of
teacher assistance during critique collection. The best result in each
column is shown in bold.}
\label{tab:difficulty_breakdown}

\begin{tabular}{
    @{}
    ll
    *{12}{S[table-format=2.1]}
    @{}
}
\toprule
\textbf{Model}
& \textbf{Training}
& \multicolumn{3}{c}{\textbf{Improve Acc.}}
& \multicolumn{3}{c}{\textbf{F1}}
& \multicolumn{3}{c}{\textbf{Direction-3}}
& \multicolumn{3}{c}{\textbf{5-Ladder}}
\\

\cmidrule(lr){3-5}
\cmidrule(lr){6-8}
\cmidrule(lr){9-11}
\cmidrule(lr){12-14}

&
& \multicolumn{1}{c}{\textbf{Sampled}}
& \multicolumn{1}{c}{\textbf{Revised}}
& \multicolumn{1}{c}{\textbf{Guided}}
& \multicolumn{1}{c}{\textbf{Sampled}}
& \multicolumn{1}{c}{\textbf{Revised}}
& \multicolumn{1}{c}{\textbf{Guided}}
& \multicolumn{1}{c}{\textbf{Sampled}}
& \multicolumn{1}{c}{\textbf{Revised}}
& \multicolumn{1}{c}{\textbf{Guided}}
& \multicolumn{1}{c}{\textbf{Sampled}}
& \multicolumn{1}{c}{\textbf{Revised}}
& \multicolumn{1}{c}{\textbf{Guided}}
\\
\midrule

Qwen-3.5-9B
& Base
& 54.2 & 50.6 & 45.0
& 54.2 & 59.2 & 54.5
& 46.8 & 42.6 & 34.6
& 34.4 & 20.0 & 12.0
\\

Qwen-3.5-35B
& Base
& 58.0 & 47.4 & 44.4
& 58.0 & 51.7 & 46.5
& 48.0 & 35.4 & 25.8
& 29.8 & 20.0 & 8.6
\\

Gemini-3.1-Pro
& Zero-shot
& 74.8 & 46.6 & 41.4
& 77.2 & 55.3 & 47.4
& \bfseries 72.4 & 39.8 & 27.6
& \bfseries 60.6 & 17.8 & 4.6
\\

\midrule

Qwen-3.5-9B
& SFT
& 72.0 & 61.2 & 54.0
& 77.6 & 71.1 & 64.2
& 65.2 & 55.2 & 46.4
& 48.0 & 27.2 & 19.4
\\

Qwen-3.5-9B
& SFT+GRPO
& 72.4 & 65.4 & 61.4
& 77.5 & 74.2 & \bfseries 72.7
& 69.0 & \bfseries 63.6 & 52.8
& 56.0 & 37.8 & 28.0
\\

Qwen-3.5-35B
& SFT
& \bfseries 75.0 & 61.6 & 60.6
& 80.1 & \bfseries 74.9 & 67.1
& 69.0 & 60.2 & 51.6
& 49.8 & 37.4 & 28.2
\\

Qwen-3.5-35B
& SFT+GRPO
& 74.0 & \bfseries 67.0 & \bfseries 64.4
& \bfseries 82.0 & 73.5 & 69.8
& 71.4 & 61.4 & \bfseries 56.0
& 57.8 & \bfseries 41.0 & \bfseries 37.8
\\

\bottomrule
\end{tabular}
\end{table*}

\section{OpenEvolve Evaluation Details}
\label{app:openevolve_details}

\subsection{OpenEvolve Framework}
\label{app:openevolve-framework}

\paragraph{Evolutionary search.}
OpenEvolve~\citep{sharma2025openevolve} is an evolutionary coding framework for recursive self-improvement. It begins with an executable seed program and maintains a population of previously evaluated solutions. At each evolutionary round, OpenEvolve samples a parent solution, together with additional high-performing or diverse solutions that serve as inspirations, and prompts a language model to propose a modification. The resulting child program is executed by a task-specific evaluator, which returns empirical performance metrics and execution feedback. The evaluated child is then added to the population and can be selected as a parent or inspiration in subsequent rounds. Repeating this process allows improvements discovered in earlier rounds to be accumulated into progressively stronger solutions.

\paragraph{Population and island management.}
OpenEvolve combines a quality-diversity archive based on MAP-Elites with an island-based population structure. MAP-Elites partitions candidate solutions according to configurable behavioral or structural features and retains a high-performing solution within each occupied region, preserving diverse approaches rather than optimizing only a single lineage. Multiple islands maintain semi-independent populations and therefore explore different evolutionary trajectories. High-performing solutions are periodically migrated between islands, allowing useful discoveries from one population to influence the others while reducing premature convergence. Parent selection balances exploitation of high-performing solutions with exploration of diverse or less frequently sampled regions of the population. OpenEvolve also tracks the best verified solution separately so that it is preserved throughout evolution.

\paragraph{Configuration in our experiments.}
For each task and screening setting, we maintain three islands and evolve each island for 10 rounds. Exactly one selected idea is implemented and empirically verified in each island round, resulting in a total budget of 30 verification runs per task. We set migration interval as 4(migration happens every 4 evolution rounds). The idea proposer is using temperature of 0.8, and idea and implementation summarizer is using temperature of 0.4. All critic models are using temperature of 0.7. All settings begin from the same seed implementation and use the same proposer, implementer, summarizer, population configuration, and verification budget; they differ only in whether and how candidate ideas are screened before execution.

\subsection{Critic-Guided Candidate Selection}\label{app:openevolve-idea-selection}

At each evolutionary round, Gemini-3.1-Pro generates a pool of 16 candidate ideas conditioned on the current parent solution. For each candidate, the critic generates 16 critiques comparing it with the parent and predicts one of five outcome labels. We rank candidates by their predicted labels in the order $\textsc{BigWin} > \textsc{MediumWin} > \textsc{SmallWin} > \textsc{Tie} > \textsc{Lose}$ and calculate an average rank for each candidate idea. Only the candidate idea with the highest average rank for execution.  Consequently, exactly one of the 16 candidates consumes a real verification run, while the remaining candidates are filtered without executing their training pipelines.

We apply the same candidate-generation and selection procedure when using Gemini-3.1-Pro, the base Qwen-3.5-9B/35B models,
or our trained Critic-9B/35B as the selector. Vanilla OpenEvolve does not apply this critic-based screening stage.

\subsection{Task Descriptions}
\label{app:mls-lite-details}

\begin{table*}[t]
\centering
\scriptsize
\setlength{\tabcolsep}{5pt}
\renewcommand{\arraystretch}{1.12}
\caption{The 30 tasks in MLS-Bench-Lite, grouped by their 12 official domains.
Tasks marked with \textsuperscript{$\dagger$} were included in the task environments used for SFT data collection, while tasks marked with\textsuperscript{$\ddagger$} were introduced only during critic model GRPO training.  Unmarked tasks were held out from critic model training. 
We retain all 30 tasks to preserve the complete MLS-Bench-Lite evaluation suite.
}
\label{tab:mls-lite-tasks}

\begin{tabularx}{\textwidth}{
    @{}
    >{\raggedright\arraybackslash}p{0.21\textwidth}
    >{\centering\arraybackslash}p{0.035\textwidth}
    >{\raggedright\arraybackslash}X
    @{}
}
\toprule
\textbf{Domain} & \textbf{\#} & \textbf{Tasks} \\
\midrule

Language Models
& 3
& \texttt{llm-dllm-demask-strategy};
  \texttt{llm-pretrain-optimizer};
  \texttt{llm-rl-importance-sampling}
\\

Robotics
& 5
& \texttt{jepa-planning};
  \texttt{robo-diffusion-guidance};
  \texttt{robo-diffusion-policy};
  \texttt{robo-humanoid-sim2real-algo};
  \texttt{robomimic-bc-loss}
\\

Vision and Generation
& 3
& \texttt{cv-3dgs-densification};
  \texttt{cv-dbm-sampler};
  \texttt{cv-vae-loss}
  \textsuperscript{$\dagger$}
\\

Reinforcement Learning
& 1
& \texttt{rl-value-discrete}
\\

ML Systems and Efficient ML
& 3
& \texttt{llm-ptq-algorithm};
  \texttt{llm-qat-algorithm};
  \texttt{mlsys-sparse-attention-inference}
\\

AI for Science
& 3
& \texttt{ai4bio-mutation-effect-prediction};
  \texttt{ai4sci-inverse-diffusion-algo};
  \texttt{ai4sci-pla-binding-affinity}
\\

Optimization and Theory
& 2
& \texttt{optimization-multi-objective};
  \texttt{optimization-variance-reduction}
  \textsuperscript{$\ddagger$}
\\

Classical and Adaptive Learning
& 2
& \texttt{ml-clustering-algorithm}
\textsuperscript{$\dagger$};
  \texttt{ml-dimensionality-reduction}
  \textsuperscript{$\dagger$}
\\

Deep Learning
& 2
& \texttt{cv-pooling-aggregation};
  \texttt{dl-activation-function}
  \textsuperscript{$\dagger$}
\\

Time Series and Forecasting
& 3
& \texttt{quant-concept-drift};
  \texttt{ts-exogenous-forecast}
  \textsuperscript{$\dagger$};
  \texttt{ts-imputation}
  \textsuperscript{$\dagger$}
\\

Structured and Causal Reasoning
& 2
& \texttt{causal-discovery-discrete}
\textsuperscript{$\ddagger$};
  \texttt{graph-generation}
  \textsuperscript{$\dagger$}
\\

Trustworthy Learning
& 1
& \texttt{security-membership-inference-defense}
\textsuperscript{$\ddagger$}
\\

\midrule
\textbf{Total}
& \textbf{30}
& \textbf{12 domains}
\\
\bottomrule
\end{tabularx}
\end{table*}

\paragraph{MLS-Bench-Lite.}
MLS-Bench~\citep{lyu2026mls} evaluates whether an ML agent can improve a targeted component of an existing machine learning system. Each task provides a fixed research scaffold, relevant source code, baseline implementations, a constrained edit surface, and an executable evaluator. The agent must propose and implement an algorithmic modification that improves performance across the task's prescribed evaluation settings. These tasks therefore test not only whether an agent can produce executable code, but also whether it can identify ML modifications that yield robust empirical improvements. We use MLS-Bench-Lite, the official 30-task subset of the full 140-task benchmark. The subset preserves coverage of all 12 MLS-Bench domains: language models, robotics, vision and generation, reinforcement learning, ML systems and efficient ML, AI for science, optimization and theory, classical and adaptive learning, deep learning, time series and forecasting, structured and causal reasoning, and trustworthy learning. The 30 tasks cover a broad range of ML development problems, including model architecture design, training objectives, optimization algorithms, inference procedures, data processing, and scientific modeling. The complete task list and domain assignments are provided in Table~\ref{tab:mls-lite-tasks}.

\paragraph{Scoring.}
Because the tasks use heterogeneous evaluation metrics, MLS-Bench defines a task-specific scoring specification for converting raw evaluation results into normalized task scores. These specifications account for metric direction, task-specific baseline anchors, theoretical or practical performance bounds, and, when applicable, multiple evaluation settings. We use the official scoring specification for each task without modification. We aggregate the normalized task scores within each domain using the arithmetic mean and report the resulting domain-level performance in Table~\ref{tab:openevolve-full}.

\paragraph{Overlap with critic model training environments.}
A subset of the MLS-Bench-Lite tasks also appears among the task environments used to collect training data for our critic models. We retain these tasks rather than removing them because our goal is to evaluate the complete official MLS-Bench-Lite suite and preserve its intended cross-domain composition. 

\subsection{OPSD Teacher vs Student Critique Reasoning Trajectory}

Table~\ref{tab:opsd-case-study} illustrates how outcome conditioning can shift a critique from dismissing a proposed modification to identifying a mechanism consistent with the observed improvement. This qualitative comparison does not independently validate the teacher's causal explanation.


\begin{table*}[t]
\centering
\small
\setlength{\tabcolsep}{6pt}
\renewcommand{\arraystretch}{1.15}
\caption{Student and outcome-conditioned teacher critiques on an
LLM PTQ example. Idea descriptions are summarized; critique
excerpts are quoted verbatim, with emphasis added.
The critiques were regenerated from an archived parent--child
pair for qualitative inspection.}
\label{tab:opsd-case-study}
\begin{tabularx}{\textwidth}{
    @{}
    >{\raggedright\arraybackslash}X
    >{\raggedright\arraybackslash}X
    @{}
}
\toprule
\multicolumn{2}{@{}p{\textwidth}@{}}{
\textbf{Task:} LLM post-training quantization
(INT4; WikiText-2 perplexity, lower is better).
} \\
\multicolumn{2}{@{}p{\textwidth}@{}}{
\textbf{Parent idea:} Apply GPTQ with Hessian-based error
compensation, dynamically computed group quantization
parameters, and damped Cholesky decomposition.
} \\
\multicolumn{2}{@{}p{\textwidth}@{}}{
\textbf{Child idea:} Integrate activation-aware channel scaling
into GPTQ, use asymmetric group quantization, and add adaptive
dampening while retaining GPTQ error compensation.
} \\
\multicolumn{2}{@{}p{\textwidth}@{}}{
\textbf{Empirical outcome:}
Perplexity decreases from $5.0428$ to $4.9906$
($1.04\%$ relative reduction); label: \textsc{Improve}.
} \\
\midrule
\textbf{Student: without outcome hint}
&
\textbf{Teacher: with outcome hint}
\\
Trained critic with the SDFT LoRA adapter enabled.
&
Frozen critic; receives the
measured parent/child metrics and outcome label.
\\
\midrule
\textbf{Quantization reasoning}
&
\textbf{Quantization reasoning}
\\
``While the CHILD introduces Asymmetric Quantization (AQ)
and AWQ scaling, AQ requires a 1-bit zero point that can
cause overflow when dequantized back to FP16.
This negates the theoretical benefit of using finer
step sizes, and the resulting perplexity remains
statistically indistinguishable from the PARENT's
symmetric GPTQ method.''
&
``By incorporating AWQ scaling into GPTQ, the CHILD
effectively reduces the quantization error variance on
salient (high-activation) channels while
\textbf{preserving the error-compensation mechanism of GPTQ}.
This leads to a lower perplexity (4.9906 vs 5.0428).''
\\
\addlinespace
\textbf{Interpretation of the proposed modification}
&
\textbf{Interpretation of the proposed modification}
\\
``The PARENT uses the full Hessian matrix for error
propagation, whereas the CHILD only uses the diagonal
(the exact same information utilized by AWQ).
The off-diagonal terms provide valuable error compensation
that the \textbf{CHILD discards}, offsetting its scaling
advantage.''
&
``The CHILD introduces a well-established and highly
effective improvement over the PARENT by
\textbf{combining AWQ's salient channel protection
with GPTQ's error compensation}.''
\\
\midrule
\textbf{Predicted outcome:} \textsc{Tie} (incorrect).
&
\textbf{Predicted outcome:} \textsc{Improve},
with a small win tier (outcome correct).
\\
\bottomrule
\end{tabularx}
\end{table*}

\section{More details on Critic-in-loop RL}

\subsection{Results for Policy Model GRPO on MLS-Lite}\label{app:policy-grpo-table}

\begin{table}[h]
\centering \small \setlength{\tabcolsep}{3.5pt} 
\caption{Policy GRPO under a fixed budget of 10,000 empirical verifications. We compare the base policy, verify-all GRPO, and critic-model-in-the-loop GRPO for Qwen-3.5-9B and Qwen-3.5-35B. Scores are averaged within each of the 11 retained MLS-Lite-Bench domains, and the best result in each row is shown in \textbf{bold}.}
\label{tab:policy-grpo-table} 
\setlength{\tabcolsep}{3pt} 
\renewcommand{\arraystretch}{1.08} 
\begin{tabularx}{\linewidth}{ @{} >{\raggedright\arraybackslash}X cccccc @{} } 
\toprule 
\multirow{2}{*}{\textbf{Domain}} & \multicolumn{3}{c}{\textbf{Qwen-3.5-9B}} & \multicolumn{3}{c}{\textbf{Qwen-3.5-35B}} \\ 
\cmidrule(lr){2-4} \cmidrule(lr){5-7} &
\textbf{Base} & \textbf{Verify-all} & \textbf{Critic-in-loop} & \textbf{Base} & \textbf{Verify-all} & \textbf{Critic-in-loop} \\

\midrule 
Language Models                  & 9.24  & 12.17 & 12.04 & 11.94 & \textbf{12.90} & 12.62 \\ 
Robotics                         & 7.91  & 18.83 & 15.87 & 16.83 & 21.20 & \textbf{23.46} \\ 
Vision and Generation            & 11.66 & 6.61  & 13.23 & 10.83 & 11.03 & \textbf{14.07} \\ 
ML Systems and Efficient ML      & 21.88 & 23.99 & \textbf{36.16} & 11.39 & 16.87 & 30.99 \\ 
AI for Science                   & 12.49 & \textbf{14.17} & 13.86 & 11.09 & 8.34  & 12.61 \\ 
Optimization and Theory          & 18.56 & 20.03 & 18.08 & 17.75 & 9.76  & \textbf{25.13} \\ 
Classical and Adaptive Learning  & 16.95 & 19.80 & 22.85 & 28.90 & \textbf{30.23} & 29.88 \\ 
Time Series and Forecasting      & 9.21  & 11.93 & \textbf{30.97} & 9.88  & 14.85 & 13.93 \\ 
Structured and Causal Reasoning  & 34.48 & 23.60 & 32.40 & 19.98 & 18.52 & \textbf{36.32} \\ 
Deep Learning                    & 41.30 & 33.05 & \textbf{48.59} & 40.37 & 41.90 & 44.50  \\
Trustworthy Learning             & 22.60 & 31.10 & \textbf{33.23} & 26.40 & 26.65 & 26.50 \\
\midrule 
Domain Average                          & 18.75 & 19.57 & \textbf{25.21} & 18.67 & 19.29 & 24.55 \\ 
\bottomrule 
\end{tabularx} 
\end{table}

\subsection{Ablation on the Critic Model as Reward Model}\label{app:ablation-critic}

We investigate whether the trained critic can support policy optimization without empirical verification during training. Starting from the base Qwen-3.5-35B policy, we perform GRPO using rewards provided entirely by the frozen Critic-35B, with no fallback to the ground-truth verifier. For each candidate idea, we aggregate $k$ critic rollouts by majority voting over \textsc{Improve}, \textsc{Tie}, and \textsc{Lose}, assigning a reward of $+1$ to \textsc{Improve} and $-1$ otherwise. We conduct two sweeps: varying the number of policy training steps from 0 to 160, and varying $k\in\{1,4,8,16,32\}$ while fixing training to 40 steps. 

\begin{figure}[h]
    \centering
    \includegraphics[width=\linewidth]{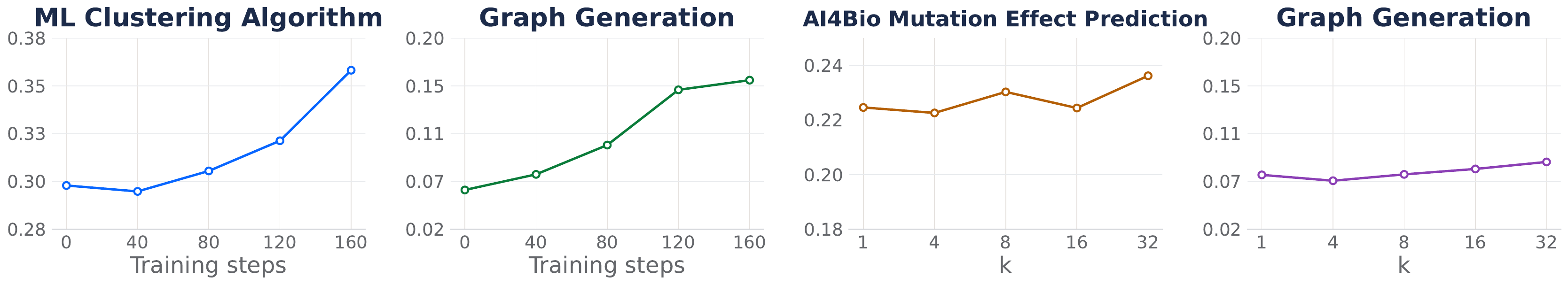}
    \caption{Critic-model-only reward ablation. The left panels vary the number of policy-training steps, while the right panels vary the number of critic model rollouts $k$ with training fixed to 40 steps.}
    \label{fig:critic-reward-ablation}
\end{figure}

Figure~\ref{fig:critic-reward-ablation} shows that policy performance generally improves as the number of training steps increases on both \texttt{ML Clustering Algorithm task} and \texttt{Graph Generation} task. The trend is particularly clear for \texttt{Graph Generation}, indicating that rewards from critic model continue to provide a useful learning signal even without access to empirical verification during training. Increasing the number of critic model rollouts $k$ also improves performance on \texttt{AI4Bio Mutation Effect Prediction} task and \texttt{Graph Generation} task, but the gains are comparatively modest and begin to saturate. In particular, increasing $k$ beyond 16 provides limited additional benefit while proportionally increasing inference computational cost. We therefore use $k=16$ as a practical trade-off between reward reliability and critic model inference computational cost. 

\end{document}